\documentclass[letterpaper]{article} 
\usepackage[]{aaai23}  
\usepackage{times}  
\usepackage{helvet}  
\usepackage{courier}  
\usepackage[hyphens]{url}  
\usepackage{graphicx} 
\usepackage{natbib}  
\usepackage{caption} 
\usepackage{algorithm}
\usepackage[usenames,dvipsnames]{color}
\usepackage{algorithmic}

\usepackage{booktabs}

\usepackage{newfloat}
\usepackage{listings}
\DeclareCaptionStyle{ruled}{labelfont=normalfont,labelsep=colon,strut=off} 
\floatstyle{ruled}
\newfloat{listing}{tb}{lst}{}
\floatname{listing}{Listing}
\title{Color Me Intrigued: Quantifying Usage of Colors in Fiction}
\author{
Siyan Li
}
\affiliations{

    Stanford University\\
    siyanli@stanford.edu
}

\begin{document}

\maketitle

\begin{abstract}
We present preliminary results in quantitative analyses of color usage in selected authors' works from LitBank. Using Glasgow Norms, human ratings on 5000+ words, we measure attributes of nouns dependent on color terms. Early results demonstrate a significant increase in noun concreteness over time. We also propose future research directions for computational literary color analytics. \footnote{All code and data used are available at \url{https://github.com/siyan-sylvia-li/ColorLit}.}
\end{abstract}

\section{Introduction}
\textit{All great writers are great colourists}, Virginia Woolf once stated \cite{woolf1934walter}. Analyzing colors in literary work across time and authors has fascinated the field of literature, philosophy, and psychology \cite{skard1946color}.

Most literary analyses of colors focus on only one author, one work, or one historical era. There has been very few large-scale analyses of color usage shifts. Recently, natural language processing (NLP) has progressed in fields potentially relevant to literary color analyses, such as dependency parsing \cite{qi2020stanza} and named entity recognition \cite{li2020survey}. Leveraging these tools expedites localization of spans of interest, increasing efficiency and ease of larger-scale literary analyses. 

What makes literary color analyses interesting for natural language processing? Authors utilize colors in numerous ways, and NLP tools should capture this variety. While Goethe uses colors as a backdrop of his narratives, only using them to emphasize the plastic shapes of objects \cite{bruns1928auge}, Dante's coloring in his \textit{Divine Comedy} displays more symbolic undertones. The colors on the three faces of Dante's Lucifer can be related to the three horses of the Apocalypse \cite{skard1946color}. The sudden absence of green in Heavenly Paradise may stem from green's association with hope, and Dante's Paradise eliminates the need for hope since it fulfills all wishes \cite{austin1933heavenly}. In contrast, James Joyce's green can be interpreted to symbolize absinthe \cite{joyce-absinthe} and the author's frustration with the Irish Catholic Church \cite{xie2015color}. For more contemporary writers, Virginia Woolf's blue in \textit{To the Lighthouse} accompanies Mrs. Ramsay for her Madonna role and her mixture of radiance and somberness \cite{stewart_color_1985}. The same blue manifests cholera and illness in Edgar Allan Poe's \textit{The Masque of the Red Death} \cite{yoon2021color}. Despite some subjectivity in these interpretations, the existence of differences in color usages are absolute. We want to examine whether current NLP tools ``understand'' these differences. 

We propose a novel line of research using word embeddings and pre-trained language models to quantify color usages in literature. Specifically, we measure the attributes of nouns dependent on color adjectives according to the Glasgow Norms \cite{scott2019glasgow}. Preliminary results demonstrate statistically significant trends over time for certain colors' Glasgow Norm attributes. We present future research directions and plausible experiments. 

Our proposed framework can supplement literary color analyses research and provide additional insight for color usages comparisons. Looking at literature and creativity through the lens of colors is informative because of the prevalence of color terms in literature. Color terms can serve as anchoring points of comparison between authors, and potentially between humans and language models. 





\section{Related Work}
The most similar work to ours would be \citet{colorblind}, a study of color term usage by both non-color-blind and color-blind individuals on Reddit. The authors discover significant differences in certain color terms. They then concentrate on the nouns that are modified by color words, using dependency parsing to obtain NOUN words in an AMOD dependency pair with an ADJ color term. The authors identify significant discrepancies between the two populations in imageability \cite{scott2019glasgow} values of color-modified nouns. Our preliminary work is methodologically similar, but we study literature instead. Additionally, our work more extensively leverages labels from the Glasgow Norms by using three dimensions instead of one.
%

Word embeddings play a crucial role in computational social science. \citet{garg2018word} leverages Word2vec \cite{mikolov2013efficient} to reflect changes in relationships between the embedding representing women and different adjectives as potentially a result of the feminist movement. A similar work, \citet{bailey2022based}, showcases that \texttt{people = man} by comparing distances between word embeddings of trait words of people, men, and women respectively. We are interested in similar techniques in a literary color analysis context.

\section{Dataset}
We use LitBank \cite{bamman2019annotated}, a Euro-centric collection of 100 English fictions from 75 authors. We conduct a scrape of Project Gutenberg using LitBank's Gutenberg ID's to obtain the full text of each work. The genres consist primarily of realistic novels, with few exceptions of science fiction (H.G. Wells, Mary Shelley), fantasy (Bram Stoker, Oscar Wilde), and horror (Edgar Allan Poe).

\section{Methodology}

\subsection{Extracting Modified Nouns}
\subsubsection{Colors and Synonyms}
We select common colors and curate a list of their synonyms. The colors include ``red", ``green", ``black", ``white", ``blue", ``brown", ``gray", ``yellow", ``pink", and ``purple". All color terms and their synonyms are in Appendix~\ref{app:color}. Each set of sentences from a Project Gutenberg E-book are split into lemmatized words. We choose sentences containing either our specified color adjectives or their synonyms for dependency parsing.

\subsubsection{Dependency Parsing}
Although \citet{colorblind} strictly studies nouns modified by color terms through the AMOD dependency, this would be limiting in lyrical writing. For instance, ``she has eyes of sapphire'' describes blue eyes and should be included in our analysis, but dependency parsing would categorize ``eyes'' and ``saphire'' to linked by NMOD instead of AMOD. Therefore, we expand upon our pool of nouns by including all nouns with a dependency link to our color terms. We employ Stanza's Dependency Parser \cite{qi2020stanza}. Upon obtaining dependencies on a sentence, we perform a filtering process to retain the relevant head-dependent pairs. The specific filtering process is as follows. For each head-dependent pair: (1) Lemmatize both the head and the dependent. (2) Iterate through all color words and their synonyms; if none of them is present in either the head or the dependent, prune out this pair. (3) If the other word in the dependency pair is not a noun or a proper noun, prune out this pair.

\begin{table}[]
    \centering
    \begin{tabular}{cc|cc}
    \toprule
        \textbf{Color} & \textbf{\# of Occurrences} & \textbf{Color} & \textbf{\# of Occurrences} \\
        \midrule
        red & 2888 & green & 1839 \\
        \midrule
        black & 3325 & white & 4990 \\
        \midrule
        blue & 1622 & brown & 1206 \\
        \midrule
        gray & 1575 & yellow & 848 \\
        \midrule
        pink & 648 & purple & 545 \\
        \bottomrule
    \end{tabular}
    \caption{The total numbers of occurrences of our selected color terms in the 100 LitBank novels. These are instances where the color terms act as either a dependent or a dependency head.}
    \label{tab:color_totals}
\end{table}

\subsection{Glasgow Norm Models}
The Glasgow Norms are a list of 5,553 words with corresponding normative human ratings on different psycholinguistic dimensions. Our ongoing work concentrates on: (1) Imageability (IMAG), the ease of summoning a mental image from a word; (2) Concreteness (CNC), the extent to which words can be experienced to our senses; and (3) Valence (VAL), how positive or negative a word's value is. We hypothesize that different authors differ on the imageability/concreteness/valence values of color-dependent nouns. 

Although the Glasgow Norms vocabulary is extensive, we still hope to handle unseen words. FastText embeddings \cite{joulin2016fasttext} reduced to 100 dimensions are used to train separate 1-layer Multi-Layer Perceptron (MLP) models to predict these values. We choose FastText for its adaptability for unseen words. Prior to training, all scores are normalized to the 0 to 1 range for better interpretability, consistent with \citet{colorblind}. Three neural networks with sigmoid activations are trained on these data, and evaluated on a held-out test set with an 8:1:1 split. We use Pearson's correlation between predictions and ground truths as our metric. \citet{colorblind} reports a Pearson's correlation of 0.76 for their IMAG model on a random held-out set, while ours achieves 0.79 on the test set. We understand that the held-out test set may be different, but this indicates that our IMAG model should as potent as the prior model. Our CNC model and VAL model achieve correlation scores of 0.83 and 0.76, respectively.

To prevent repeated occurrences of a word affecting the average Glasgow Norm values, the dependent nouns are deduplicated when computing the averages.

\section{Preliminary Results}
Despite our analyses on LitBank yield statistically significant results when analyzed across time, this could stem from an imbalance in the distribution of publication time in LitBank. This paper aims to establish a preliminary framework for studying color usages in literature, and current results would need corroboration from additional texts from different eras and genres.

\subsection{Color-dependent Nouns}

We conduct both quantitative and qualitative analyses on color-dependent nouns in LitBank through both computing average Glasgow Norm values and through inspection of most frequently associated nouns. Additional analyses of color term frequencies are in Appendix~\ref{app:freqs}.

\begin{figure}[!h]
    \centering
    \includegraphics[scale=0.25]{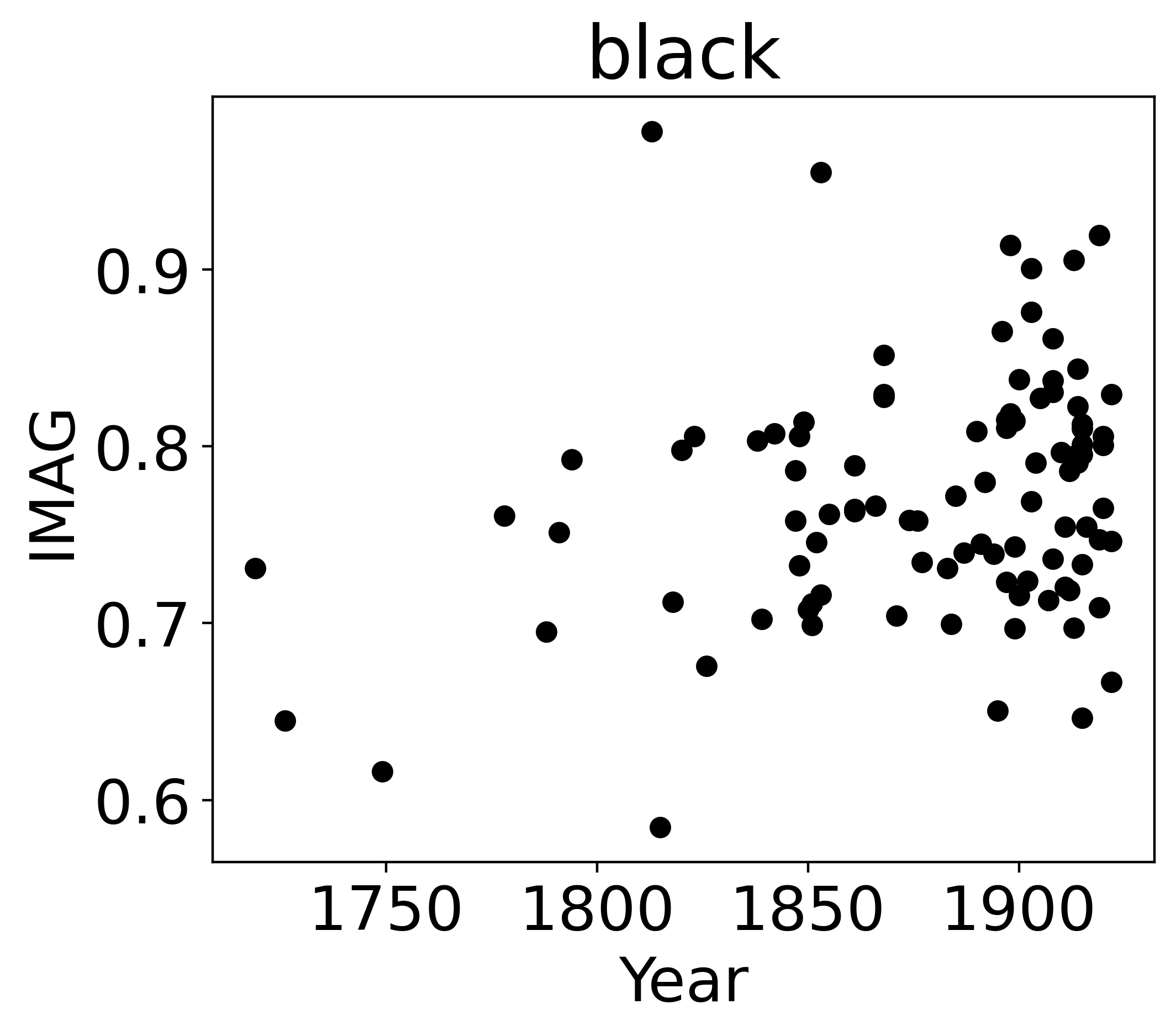}
    \includegraphics[scale=0.25]{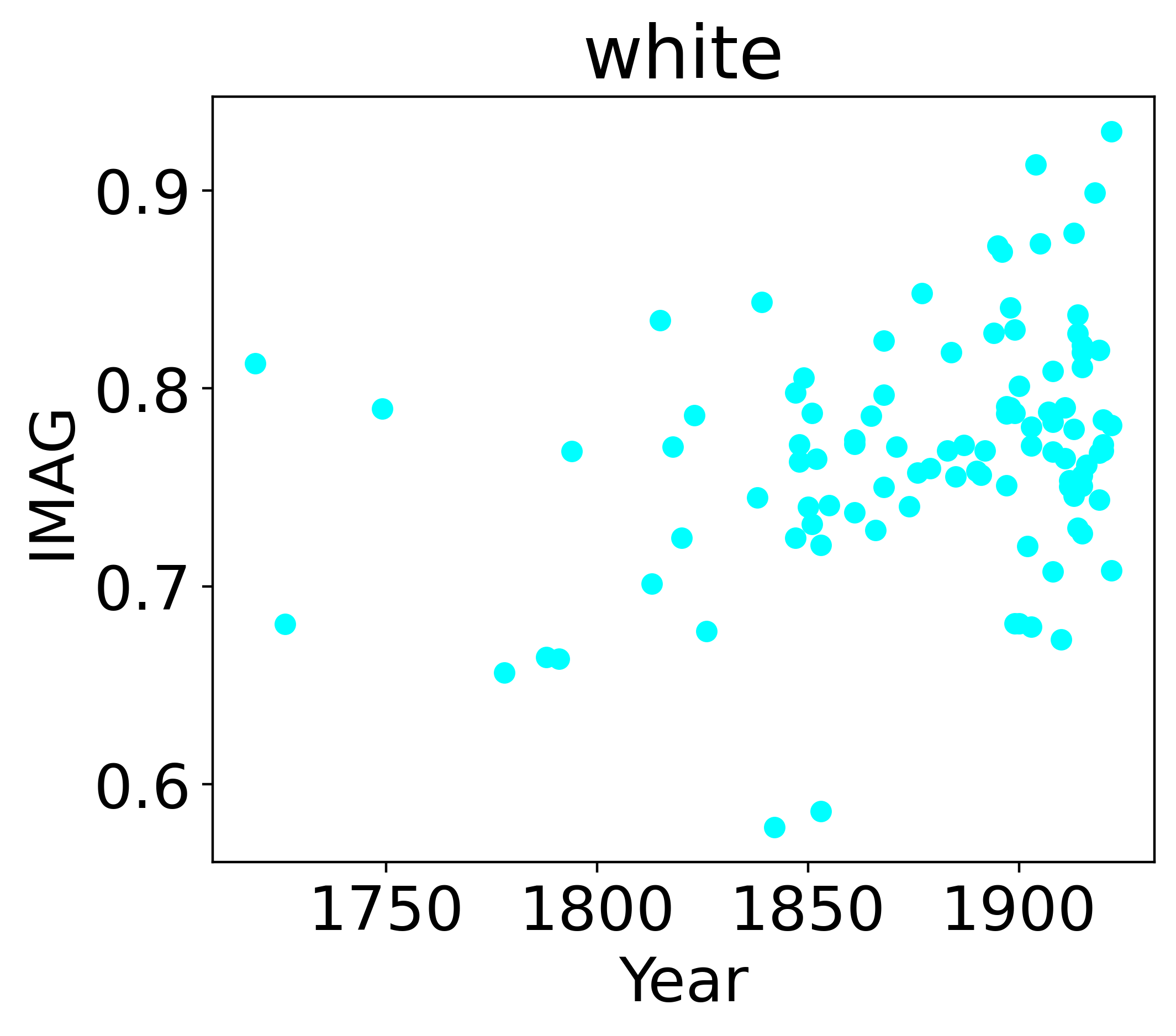}
    \includegraphics[scale=0.25]{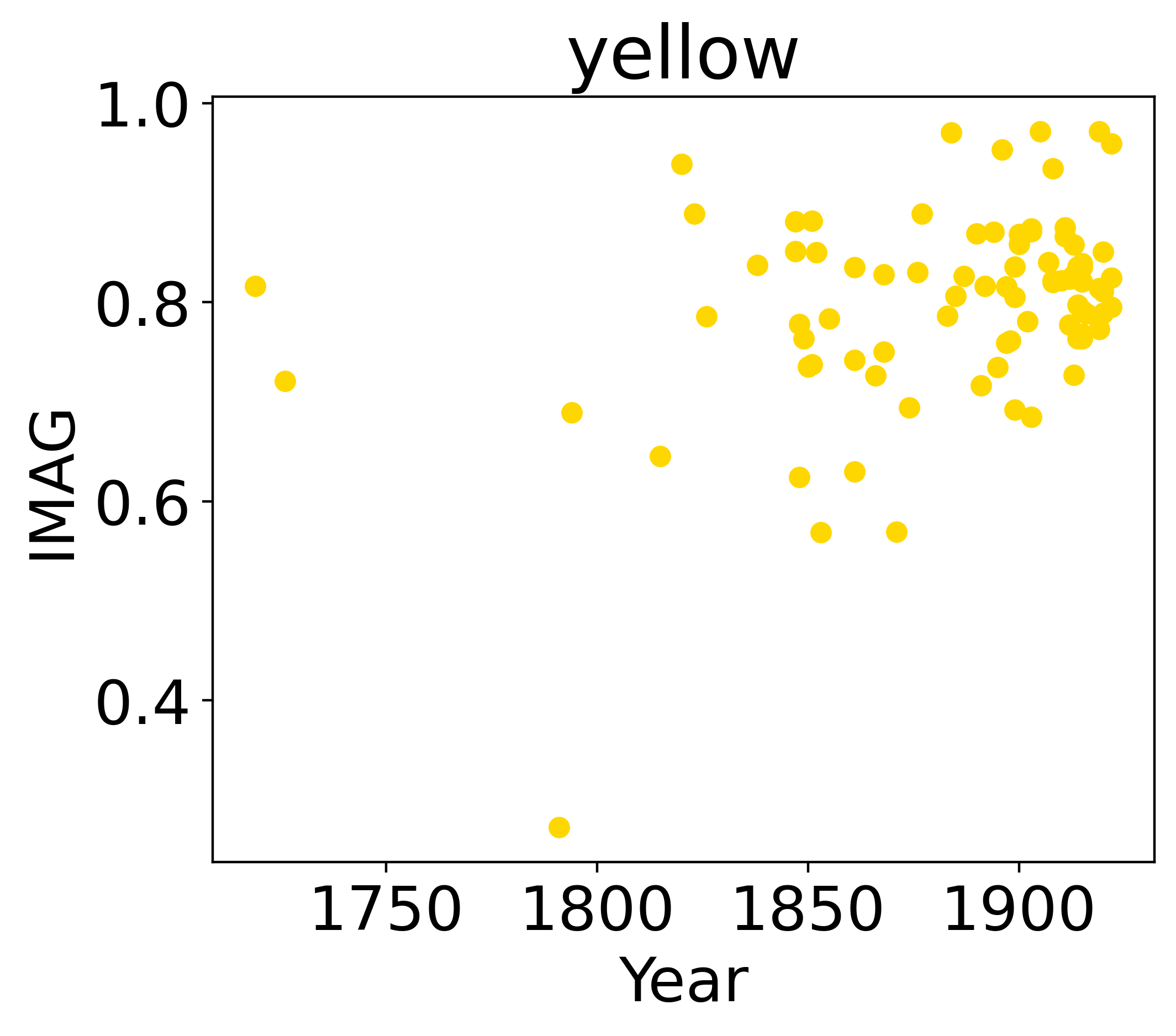}
    \includegraphics[scale=0.25]{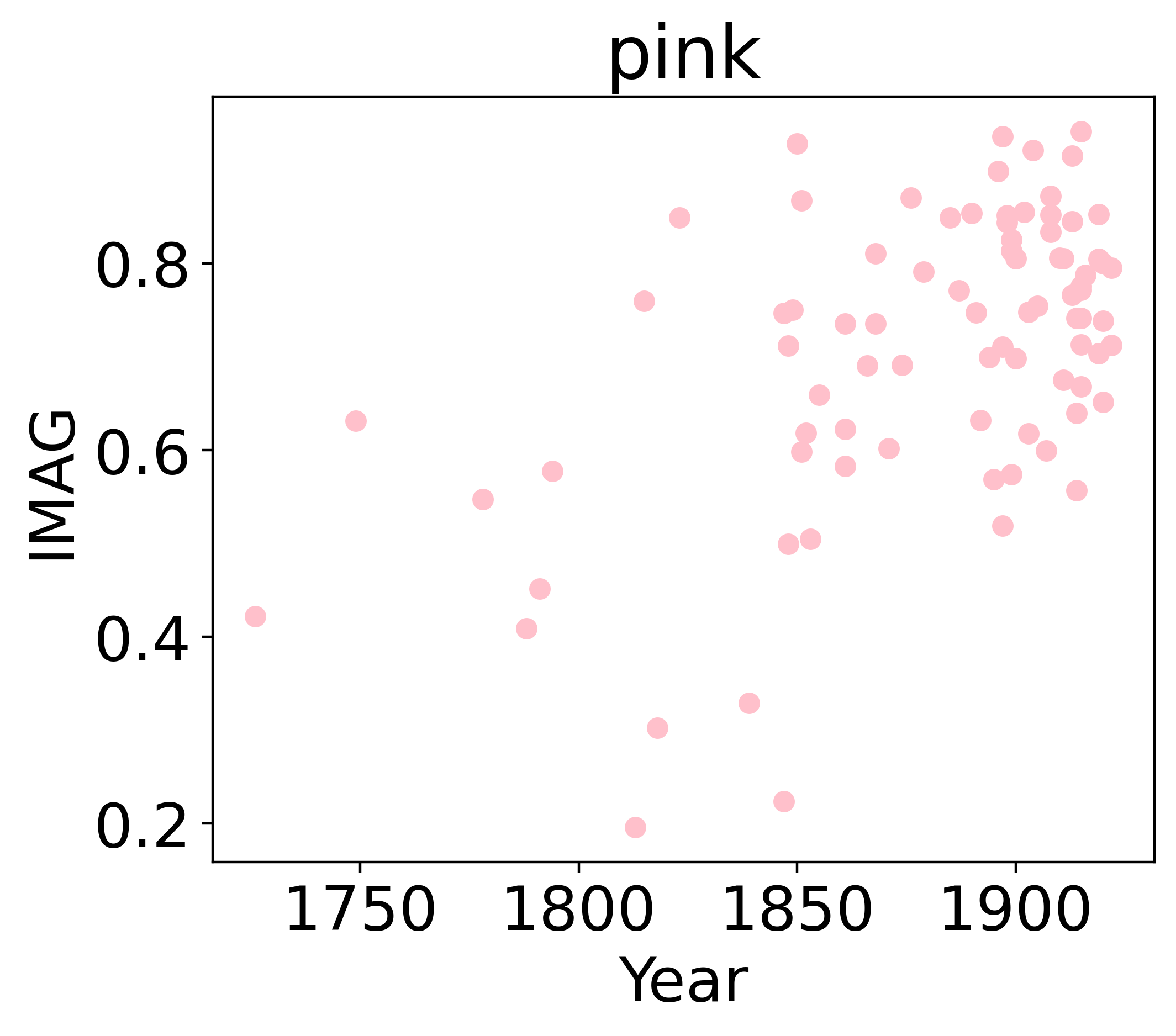}
    \caption{Imageability plots for the color terms with significant results. We omit the concreteness plots here because IMAG and CNC are highly correlated, but we provide the full set of plots in Appendix \ref{app:norms}.}
    \label{fig:imag_plots}
\end{figure}

Out of all unique nouns, 1299 are within the Glasgow Norm vocabulary, and 1924 are OOV. We use our trained MLPs to infer Glasgow Norm scales of the out-of-vocabulary nouns. After recognizing an upward trend in IMAG and CNC, we compute Pearson's correlations between publication year and average IMAG and CNC values for all color terms in novels where the color is present (Table~\ref{tab:color_pearsons}, Figure~\ref{fig:imag_plots}). The IMG and CNC values increase significantly over time for black, white, yellow, and pink. This indicates that the nouns associated with these color terms become increasingly concrete easier to conjure mental images of. This is consistent with some conclusions from \citet{skard1946color} that early color usages are obscure and abstract. 

\begin{table}[]
    \centering
    \begin{center}
        \textbf{IMAG Results}
    \end{center}
    \begin{tabular}{cc|cc}
    \toprule
        \textbf{Color} & \textbf{Pearson's r} & \textbf{Color} & \textbf{Pearson's r} \\
        \midrule
        red & -0.095* & green & 0.059 \\
        \midrule
        black & 0.257** & white & 0.303*** \\
        \midrule
        pink & 0.534*** & yellow & 0.340** \\
        \bottomrule
    \end{tabular}
    \begin{center}
        \textbf{CNC Results}
    \end{center}
    \begin{tabular}{cc|cc}
    \toprule
        \textbf{Color} & \textbf{Pearson's r} & \textbf{Color} & \textbf{Pearson's r} \\
        \midrule
        black & 0.237** & white & 0.239** \\
        \midrule
        pink & 0.517*** & yellow & 0.318*** \\
        \bottomrule
    \end{tabular}
    \begin{center}
        \textbf{VAL Results}
    \end{center}
    \begin{tabular}{cc|cc}
    \toprule
        \textbf{Color} & \textbf{Pearson's r} & \textbf{Color} & \textbf{Pearson's r} \\
        \midrule
        green & 0.273*** & purple & 0.210*\\
        \bottomrule
    \end{tabular}
    \caption{Pearson's correlations between published years and Glasgow Norm values of the 100 fictions for color terms with sig. results. *** $p < 0.001$, ** $p < 0.05$, and * $p < 0.1$. Full results are in Appendix~\ref{app:norms}.}
    \label{tab:color_pearsons}
\end{table}

\subsubsection{Nouns in Individual Works}
We observe interesting data points in our plots (the full sets of figures are available in Appendix~\ref{app:norms}). For instance, we notice a significantly lower valence for red in Edgar Allan Poe's \textit{The Masque of the Red Death}, because the only nouns associated with red are ``death'', ``stain'', and ``horror''. Similarly, an abnormally low green valence arises in Henry Fielding's \textit{History of Tom Jones, a Foundling}, because green is attached to ``slut'', ``witch'', and ``monster''.

\subsubsection{Nouns over Publication Time}
We divide our fictions into pre-1800's, 1800 - 1900, and post-1900 based on patterns in our data. We then deduplicate dependent nouns in each work so that we can measure most frequent nouns in each era without thematic motifs biasing our analyses. A list of selected color terms and their most dependent nouns in each of the eras are in Table~\ref{tab:nouns}.

\begin{table}[!h]
    \centering
    \begin{tabular}{lcc}
    \toprule
       \textbf{Color} & \textbf{Era} & \textbf{Frequent Nouns}  \\
       \midrule
        pink & Pre-1800 & \begin{tabular}[c]{@{}c@{}}shame, guilt, folly, \\ribbon, indignation\end{tabular} \\
        & 1800 - 1900 & cheek, face, ribbon, rose, lip \\
        & Post-1900 & cheek, face, rose, paper, bud \\
        \midrule
        black & Pre-1800 & color, eye, grain, hair, wave \\
        & 1800 - 1900 & hair, eye, shadow, dress, face \\
        & Post-1900 & hair, eye, dress, figure, man \\
        \midrule
        white & Pre-1800 &  \begin{tabular}[c]{@{}c@{}}face, cheek, countenance, \\cliff, cave \end{tabular} \\
        & 1800 - 1900 & face, cheek, hand, hair, man \\
        & Post-1900 & face, hand, eye, hair, man \\
        \midrule
        yellow & Pre-1800 & \begin{tabular}[c]{@{}c@{}}appearance, complexion, \\blossom, lustre, mist\end{tabular} \\
        & 1800 - 1900 & hair, light, face, glove, skin \\
        & Post-1900 & hair, light, eye, flower, house \\
        \bottomrule
    \end{tabular}
    \caption{Most frequently color-modified nouns from each era for selected color terms. Example sentences are in Appendix~\ref{app:examples}}
    \label{tab:nouns}
\end{table}

From the table we observe a significant shift in frequently used nouns across time. Pre-1800 dependent nouns are more abstract and complex compared to post-1800 nouns dependent on the same colors, while there is no significant difference between 1800 - 1900 and post-1900. This is a possible explanation for the increase in imageability and concreteness over time among LitBank works.



\subsection{Inter-Author Differences}
We plot the Word2vec embeddings of nouns dependent on the same colors from different authors to decipher how the color terms are used. Out-of-vocabulary nouns are discarded. This serves as a crude visualization of different topics associated with these color terms. For instance, when comparing nouns modified by \textit{yellow} in works of Fitzgerald and Joyce in LitBank, the topic of facial hair (hair, beard, pompadour) only manifests in Fitzgerald's, while food items (soup, cheese) appear in Joyce's. Additional examples are in Appendix~\ref{app:examples}.

\section{Proposal of Future Work}
\subsection{Further Analyses}
\subsubsection{Fine-grained Timeline Analyses}
Similar to \citet{garg2018word}, we can train separate Word2vec models on each decade of literature in our collection for fine-grained analyses. Preliminary results indicate that certain colors become increasingly associated with concrete descriptions (pink associated with cheeks and face). We can compute cosine similarities between Word2vec embeddings of certain colors with words such as ``face'' and ``lips''; this should increase over time as we observe increasing presence of colors in character descriptions. A similar metric can further quantize inter-author differences as well.

\subsubsection{Additional Clustering}
We demonstrate the prowess of Word2vec for visualizing color-related topics in this proposal, but word embeddings fail to account for contexts. Further clustering analyses can include embeddings from context-dependent pre-trained language models such as SentenceBERT \cite{reimers2019sentence} and BERT \cite{devlin2018bert}.

\begin{figure}
    \centering
    \includegraphics[scale=0.3]{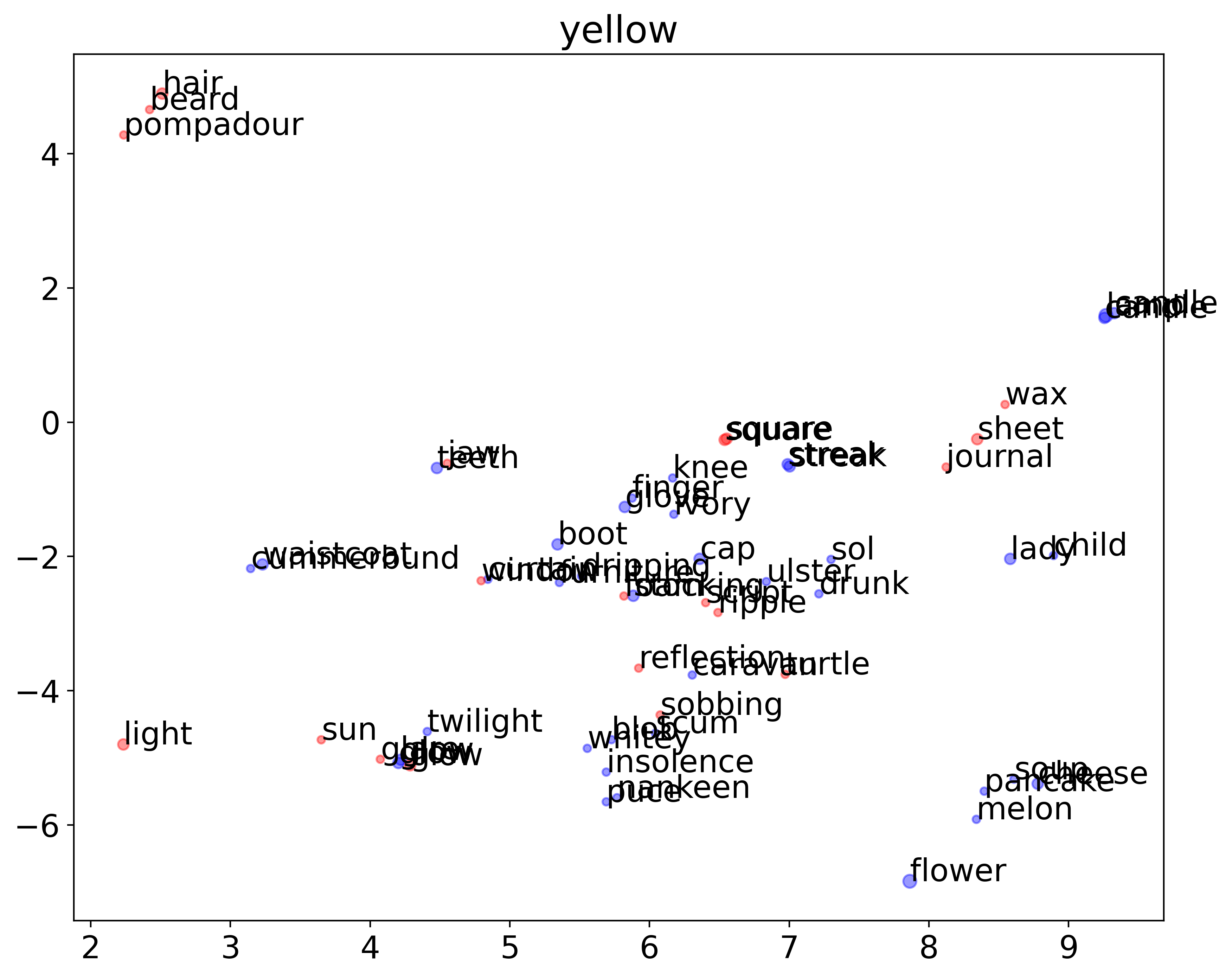}
    \caption{Word2vec embeddings of nouns modified by \textit{yellow} in novels by F. Scott Fitzgerald (red points) and James Joyce (blue points).}
    \label{fig:joyce_fitz}
\end{figure}

\subsection{RQs: From a Literature Perspective}
\subsubsection{How do colors differ across genres?}
We chose LitBank for its ease of access and thorough documentations, but one emerging issue is that LitBank skews heavily towards realistic fictions. Due to this imbalance, we cannot compare color usages meaningfully across genres. We will include more books in future analyses. If we observe significant differences in frequencies of different color words or in the concepts and objects associated with the colors, we can conclude that there exists cross-genre differences in color usage.

We already observe such differences. Bram Stoker's \textit{Dracula} \cite{stoker1897dracula}, a pioneering work in vampire literature, features numerous descriptions of pale maidens with their red lips, as well as of scarlet blood and crimson eyes. Therefore, we notice much more frequent usage of the color red in this work, compared to H. G. Wells's \textit{The War of the Worlds} \cite{wells2003war}, where red most commonly modifies weeds. We want to inspect whether the same general pattern would persist on larger-scale analyses.

\subsubsection{How do colors differ across literary forms?}
Literary color analyses often separate novelists and poets. Comparisons are often drawn only between two poets or two novelists, but rarely both. Our hypothesis is that colors usage in poetry differs significantly from color usage in prose and novels. Such differences can manifest as a discrepancy in concreteness (e.g. in poetry, colors can more often associate with a concept instead of a concrete object). Expanding our research to include poets, preferably those contemporary to our novelists, should enable us to address this systematically.

\subsection{RQs: From a Social Science Perspective}
\subsubsection{How do colors differ across cultures and classes?} 
Different cultures can have different associations with the same color; for instance, white is often a symbol of purity and a staple at Western weddings, whereas the same color is used more traditionally in Chinese funerals. These associations may reflect socio-economic classes as well (e.g. white-collar and blue-collar jobs) through colors frequently co-occurring with characters from different cultures and societal classes. To analyze this, we will utilize named entity recognition to link colors to characters, operating with more context. Pursuing this research direction would involve translated work, instead of using only Euro-centric collections of literature. The social class of a character can either be looked up online or inferred by a model to automate the process. We can then cluster character colors by cultures and classes.

\subsubsection{How do colors affect biases and stereotypes?}
Current-day color associations, such as pink with girls and blue with boys, can fuel biases. For instance, boys who enjoy wearing pink may be regarded as ``girly'' and overly feminine. Certain colors also contain associations with LGBTQ+ communities. We are interested in identifying how these color-based biases (going beyond race) manifest in literature and online communities. We can study this by finding colors associated with characters of a demographic group of interest. Tracing through past literature may also shed light upon the evolution of color associations. 

\section{Discussion}
Our work serves as a step towards more systematic analyses of color usages in literature using natural language processing tools. Following prior work, we propose using The Glasgow Norms and word embeddings as tools for quantifying color usage differences. We demonstrate significant increasing trends in imageability and concreteness in color-dependent nouns over time.

One limitation is the range of language we are capable of handling is constrained by the language models we employ. Our current collection does not have many pre-1800 pieces. While it is possible to increase the representation of pre-1800s literature, the domain shift in English style and word conventions may require different word embedding and pre-trained models to embed narratives (e.g. Chaucer often uses ``red'' in place of ``read'', standard of his time, but causes ambiguity when processing texts on a large scale). Given such shifts in vocabulary and sentence structures, we may fail to provide meaningful insights into earlier literature, since the word embeddings may have disjoint vocabularies, and models such as SentenceBERT are trained on more modern texts.

\bibliography{aaai23}

\begin{thebibliography}{21}
\providecommand{\natexlab}[1]{#1}

\bibitem[{Austin(1933)}]{austin1933heavenly}
Austin, H.~D. 1933.
\newblock Heavenly Gold; A Study of the Use of Color in Dante.
\newblock \emph{Philological Quarterly}, 12: 44.

\bibitem[{Bailey, Williams, and Cimpian(2022)}]{bailey2022based}
Bailey, A.~H.; Williams, A.; and Cimpian, A. 2022.
\newblock Based on billions of words on the internet, people= men.
\newblock \emph{Science Advances}, 8(13): eabm2463.

\bibitem[{Bamman, Popat, and Shen(2019)}]{bamman2019annotated}
Bamman, D.; Popat, S.; and Shen, S. 2019.
\newblock An annotated dataset of literary entities.
\newblock In \emph{Proceedings of the 2019 Conference of the North American
  Chapter of the Association for Computational Linguistics: Human Language
  Technologies, Volume 1 (Long and Short Papers)}, 2138--2144.

\bibitem[{Bruns(1928)}]{bruns1928auge}
Bruns, F. 1928.
\newblock Auge und Ohr in Goethes Lyrik.
\newblock \emph{The Journal of English and Germanic Philology}, 27(3):
  325--360.

\bibitem[{Devlin et~al.(2018)Devlin, Chang, Lee, and
  Toutanova}]{devlin2018bert}
Devlin, J.; Chang, M.-W.; Lee, K.; and Toutanova, K. 2018.
\newblock Bert: Pre-training of deep bidirectional transformers for language
  understanding.
\newblock \emph{arXiv preprint arXiv:1810.04805}.

\bibitem[{Earle(2003)}]{joyce-absinthe}
Earle, D.~M. 2003.
\newblock "Green Eyes, I See You. Fang, I Feel": The Symbol of Absinthe in
  "Ulysses".
\newblock \emph{James Joyce Quarterly}, 40(4): 691--709.

\bibitem[{Garg et~al.(2018)Garg, Schiebinger, Jurafsky, and Zou}]{garg2018word}
Garg, N.; Schiebinger, L.; Jurafsky, D.; and Zou, J. 2018.
\newblock Word embeddings quantify 100 years of gender and ethnic stereotypes.
\newblock \emph{Proceedings of the National Academy of Sciences}, 115(16):
  E3635--E3644.

\bibitem[{Joulin et~al.(2016)Joulin, Grave, Bojanowski, Douze, J{\'e}gou, and
  Mikolov}]{joulin2016fasttext}
Joulin, A.; Grave, E.; Bojanowski, P.; Douze, M.; J{\'e}gou, H.; and Mikolov,
  T. 2016.
\newblock FastText.zip: Compressing text classification models.
\newblock \emph{arXiv preprint arXiv:1612.03651}.

\bibitem[{Li et~al.(2020)Li, Sun, Han, and Li}]{li2020survey}
Li, J.; Sun, A.; Han, J.; and Li, C. 2020.
\newblock A survey on deep learning for named entity recognition.
\newblock \emph{IEEE Transactions on Knowledge and Data Engineering}, 34(1):
  50--70.

\bibitem[{Mikolov et~al.(2013)Mikolov, Chen, Corrado, and
  Dean}]{mikolov2013efficient}
Mikolov, T.; Chen, K.; Corrado, G.; and Dean, J. 2013.
\newblock Efficient estimation of word representations in vector space.
\newblock \emph{arXiv preprint arXiv:1301.3781}.

\bibitem[{Qi et~al.(2020)Qi, Zhang, Zhang, Bolton, and Manning}]{qi2020stanza}
Qi, P.; Zhang, Y.; Zhang, Y.; Bolton, J.; and Manning, C.~D. 2020.
\newblock Stanza: A Python natural language processing toolkit for many human
  languages.
\newblock \emph{arXiv preprint arXiv:2003.07082}.

\bibitem[{Rabinovich and Carmeli(2022)}]{colorblind}
Rabinovich, E.; and Carmeli, B. 2022.
\newblock Exploration of the Usage of Color Terms by Color-blind Participants
  in Online Discussion Platforms.
\newblock \emph{arXiv preprint arXiv:2210.11905}.

\bibitem[{Reimers and Gurevych(2019)}]{reimers2019sentence}
Reimers, N.; and Gurevych, I. 2019.
\newblock Sentence-bert: Sentence embeddings using siamese bert-networks.
\newblock \emph{arXiv preprint arXiv:1908.10084}.

\bibitem[{Scott et~al.(2019)Scott, Keitel, Becirspahic, Yao, and
  Sereno}]{scott2019glasgow}
Scott, G.~G.; Keitel, A.; Becirspahic, M.; Yao, B.; and Sereno, S.~C. 2019.
\newblock The Glasgow Norms: Ratings of 5,500 words on nine scales.
\newblock \emph{Behavior research methods}, 51(3): 1258--1270.

\bibitem[{Skard(1946)}]{skard1946color}
Skard, S. 1946.
\newblock The Use of Color in Literature: A Survey of Research.
\newblock \emph{Proceedings of the American Philosophical Society}, 90(3):
  163--249.

\bibitem[{Stewart(1985)}]{stewart_color_1985}
Stewart, J.~F. 1985.
\newblock Color in {To} the {Lighthouse}.
\newblock \emph{Twentieth Century Literature}, 31(4): 438--458.
\newblock Publisher: [Duke University Press, Hofstra University].

\bibitem[{Stoker(1897)}]{stoker1897dracula}
Stoker, B. 1897.
\newblock Dracula: 1897.

\bibitem[{Wells(1897)}]{wells2003war}
Wells, H.~G. 1897.
\newblock \emph{The war of the worlds}.
\newblock Broadview Press.

\bibitem[{Woolf(1934)}]{woolf1934walter}
Woolf, V. 1934.
\newblock \emph{Walter Sickert: A Conversation}.
\newblock L. and Virginia Woolf at the Hogarth Press.

\bibitem[{Xie(2015)}]{xie2015color}
Xie, Y. 2015.
\newblock Color as Metaphor-A Study of Joyce's Use of" Black" and" Green" in
  Dubliners and A Portrait of the Artist as a Young Man.
\newblock \emph{English Language and Literature Studies}, 5(4): 61.

\bibitem[{Yoon(2021)}]{yoon2021color}
Yoon, S. 2021.
\newblock Color Symbolisms of Diseases: Edgar Allan Poe’s “The Masque of
  the Red Death”.
\newblock \emph{The Explicator}, 79(1-2): 21--24.

\end{thebibliography}
\appendix
\section{Color terms and their synonyms used} \label{app:color}
We list the color terms used in this work, along with their corresponding synonyms obtained through manual inspection on Thesaurus.com entries, in Table~\ref{tab:color_tab}.
\begin{table}[]
    \centering
    \begin{tabular}{cc}
    \toprule
        \textbf{Color Word} & \textbf{Synonyms} \\
    \midrule
       red  & \begin{tabular}[c]{@{}l@{}}cardinal, coral, crimson, maroon, \\ burgundy, flaming, scarlet, fuchsia \end{tabular} \\
       \midrule
       green  & \begin{tabular}[c]{@{}l@{}}emerald, olive, aquamarine, beryl,\\ jade, lime \end{tabular} \\
       \midrule
       black & \begin{tabular}[c]{@{}l@{}}ebony, jet, obsidian, onyx, inky \end{tabular} \\
       \midrule
       white & \begin{tabular}[c]{@{}l@{}}alabaster, ashen, blanched, bleached, \\ cadaverous, doughy, pale, pallid, \\ pasty, ivory, pearly, beige \end{tabular} \\
       \midrule
       blue & \begin{tabular}[c]{@{}l@{}}azure, indigo, sapphire, cerulean, \\cobalt, turquoise, teal \end{tabular} \\
       \midrule
       brown & \begin{tabular}[c]{@{}l@{}}amber, khaki, tan, umber, hazel \end{tabular}\\
       \midrule
       gray & grey \\
       \midrule
       yellow & N/A \\
       \midrule
       pink & \begin{tabular}[c]{@{}l@{}}rosy, blush, magenta \end{tabular}\\
       \midrule
       purple & \begin{tabular}[c]{@{}l@{}}lavender, lilac, mauve, periwinkle, \\plum, violet, amethyst \end{tabular}\\
    \bottomrule
    \end{tabular}
    \caption{List of color words used and their synonyms}
    \label{tab:color_tab}
\end{table}

\section{Normalized Frequencies Over Time} \label{app:freqs}
After filtering through sentences in LitBank for relevant dependence pairs, the absolute frequencies of each color term are listed in Table~\ref{tab:color_totals}. We also calculate the normalized frequencies (absolute frequency divided by the total number of words) for the 100 fictions. We notice an increasing trend in the normalized frequencies over time for LitBank, and compute Pearson's correlations between normalized frequencies of color terms and time to verify this trend (Table~\ref{tab:color_freqs}).

\begin{table}[]
    \centering
    \begin{tabular}{cc|cc}
    \toprule
        \textbf{Color} & \textbf{Pearson's r} & \textbf{Color} & \textbf{Pearson's r} \\
        \midrule
        red & 0.173* & green & 0.134 \\
        \midrule
        black & 0.139 & white & 0.168* \\
        \midrule
        blue & 0.148 & brown & 0.202** \\
        \midrule
        gray & 0.162 & yellow & 0.184* \\
        \midrule
        pink & 0.173* & purple & 0.150 \\
        \bottomrule
    \end{tabular}
    \caption{Pearson's correlations between published years and normalized frequencies of the 100 fictions for each color term. ** indicates $p < 0.05$, and * $p < 0.1$.}
    \label{tab:color_freqs}
\end{table}

\begin{figure*}
    \centering
    \includegraphics[scale=0.4]{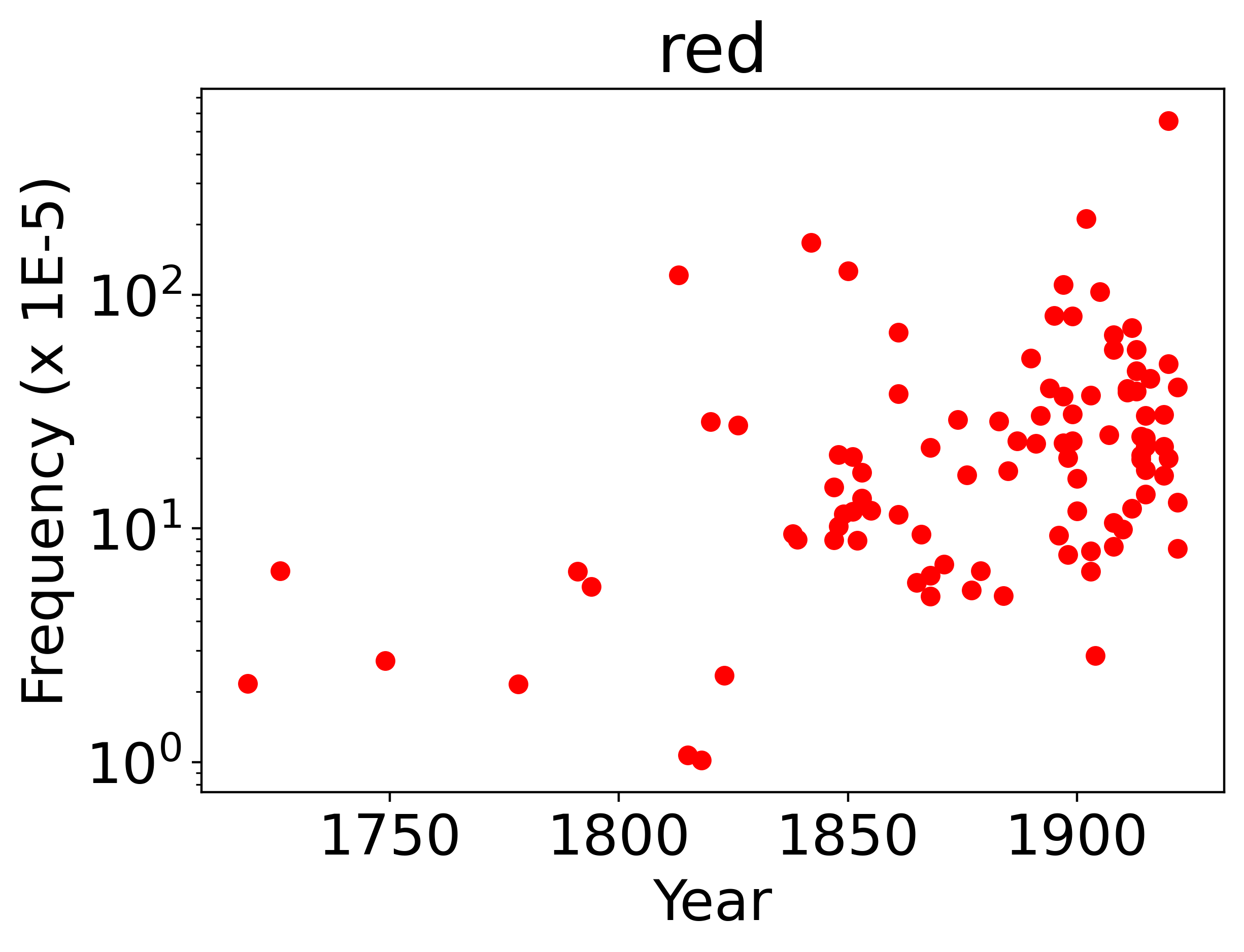}
    \includegraphics[scale=0.4]{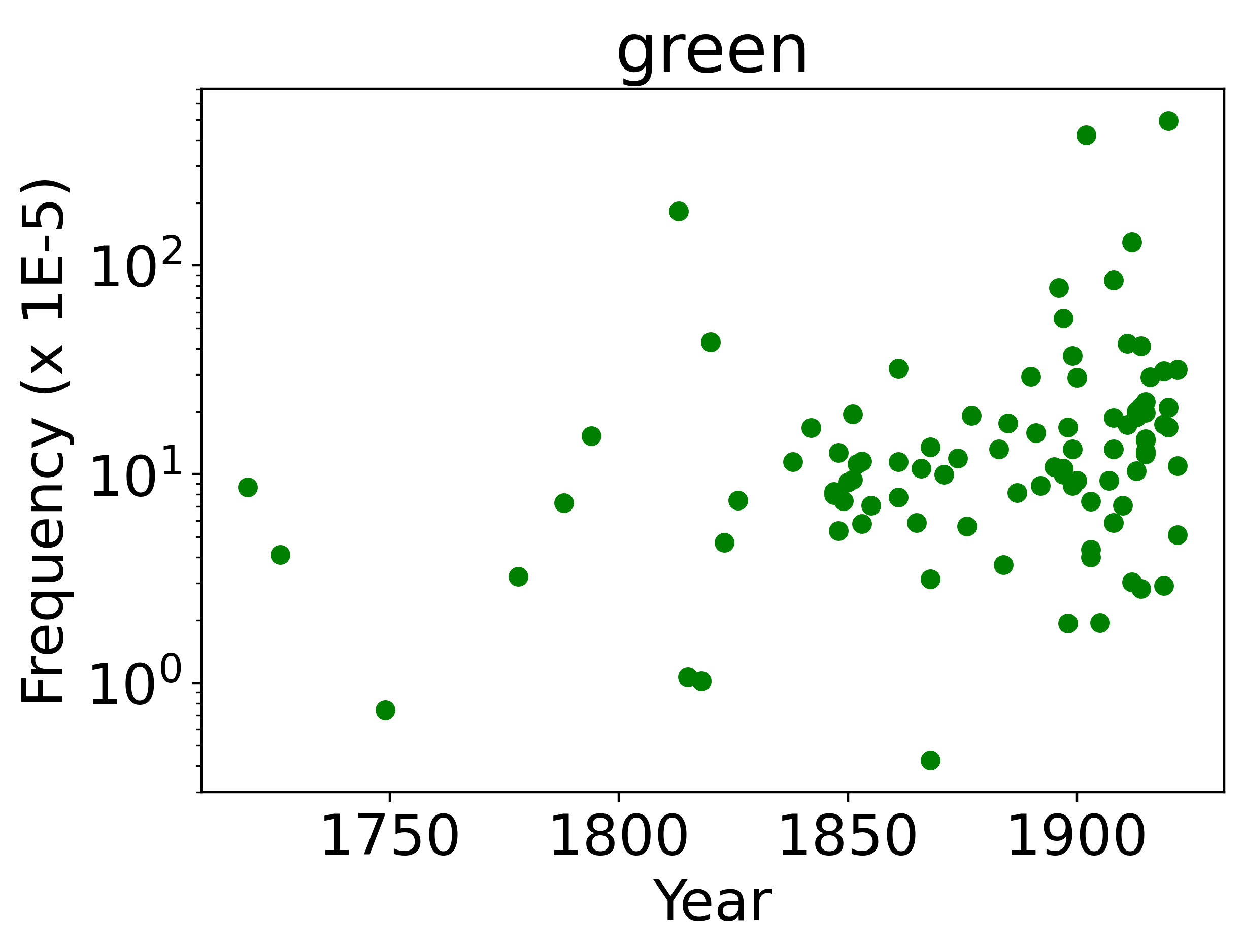}
    \includegraphics[scale=0.4]{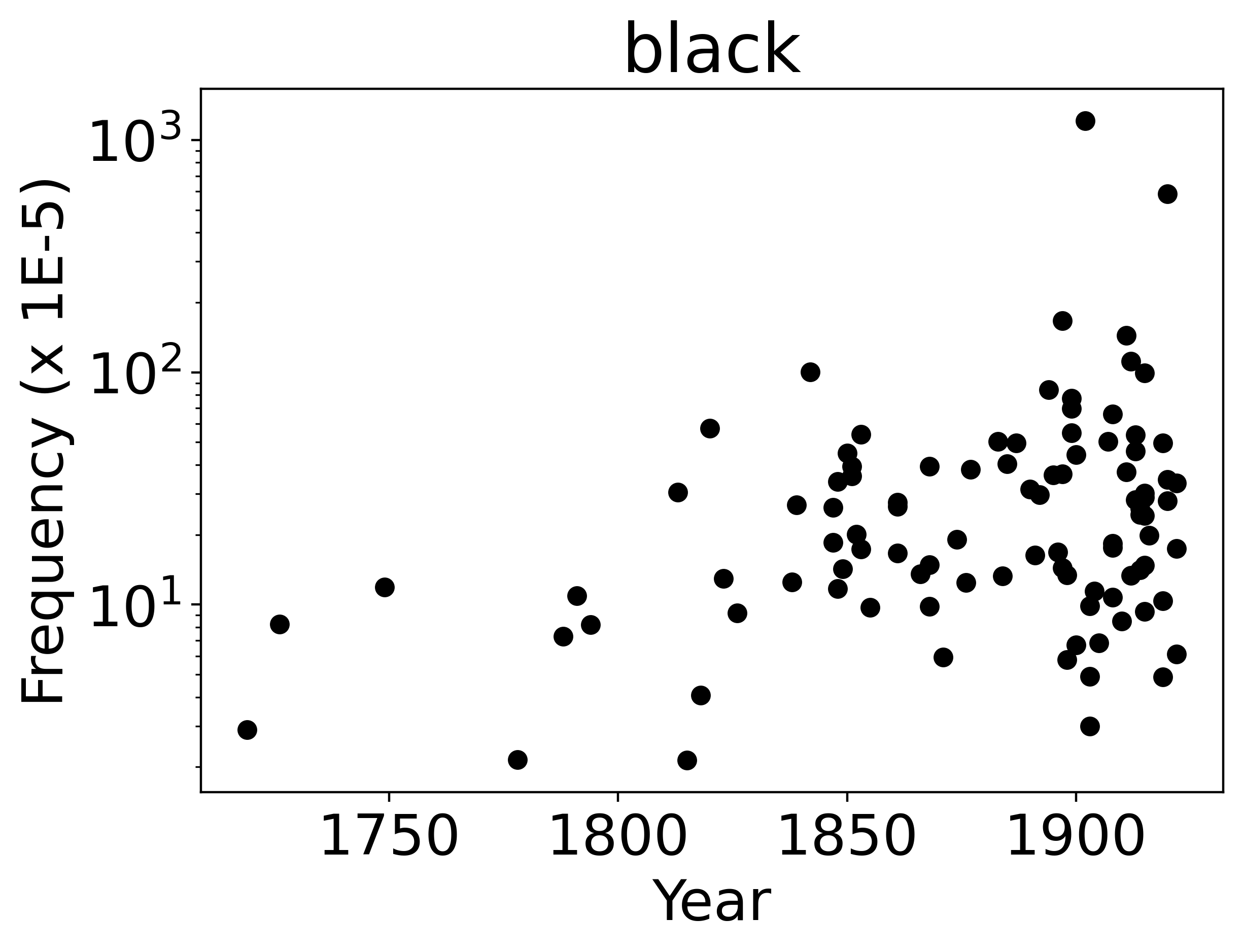}
    \includegraphics[scale=0.4]{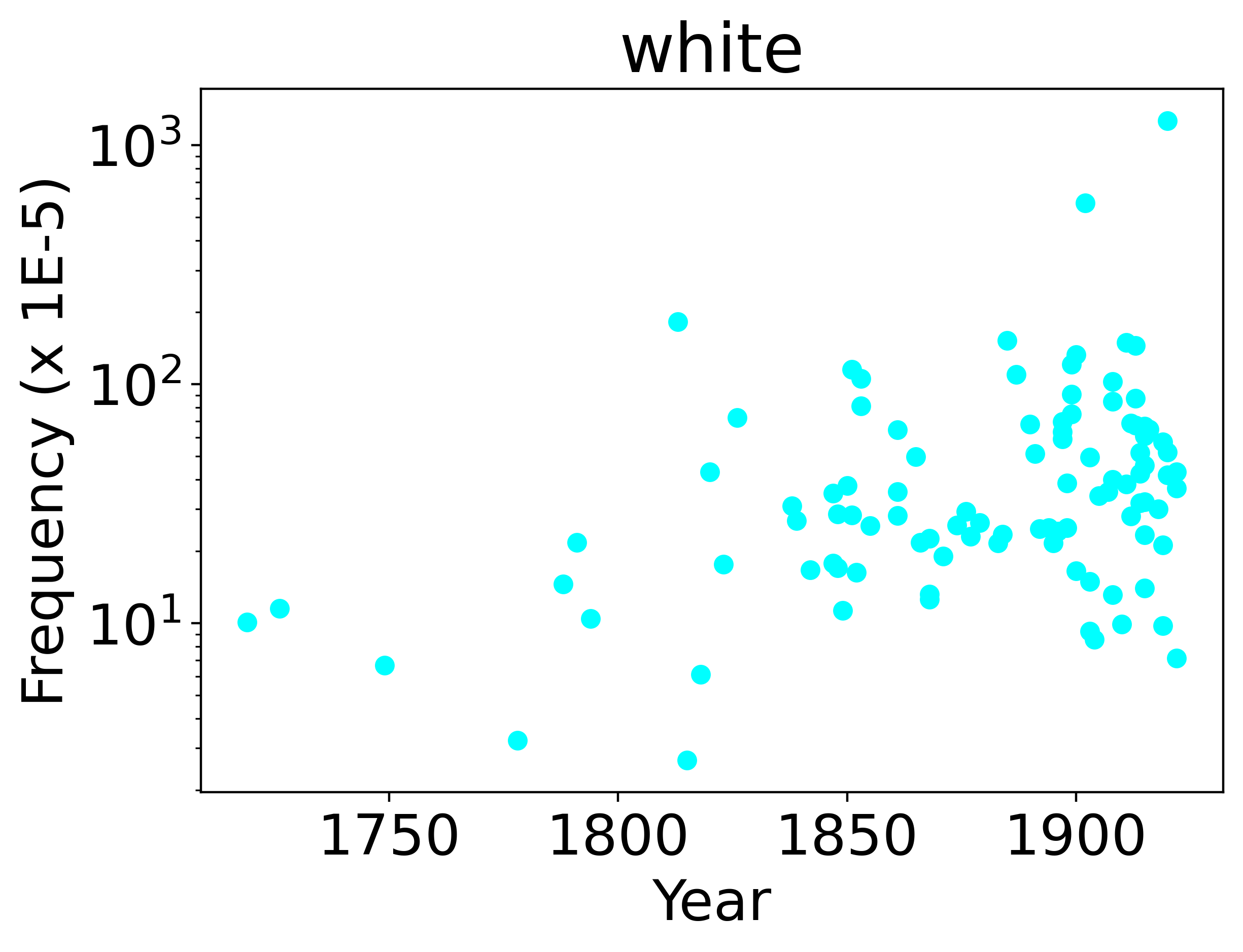}
    \includegraphics[scale=0.4]{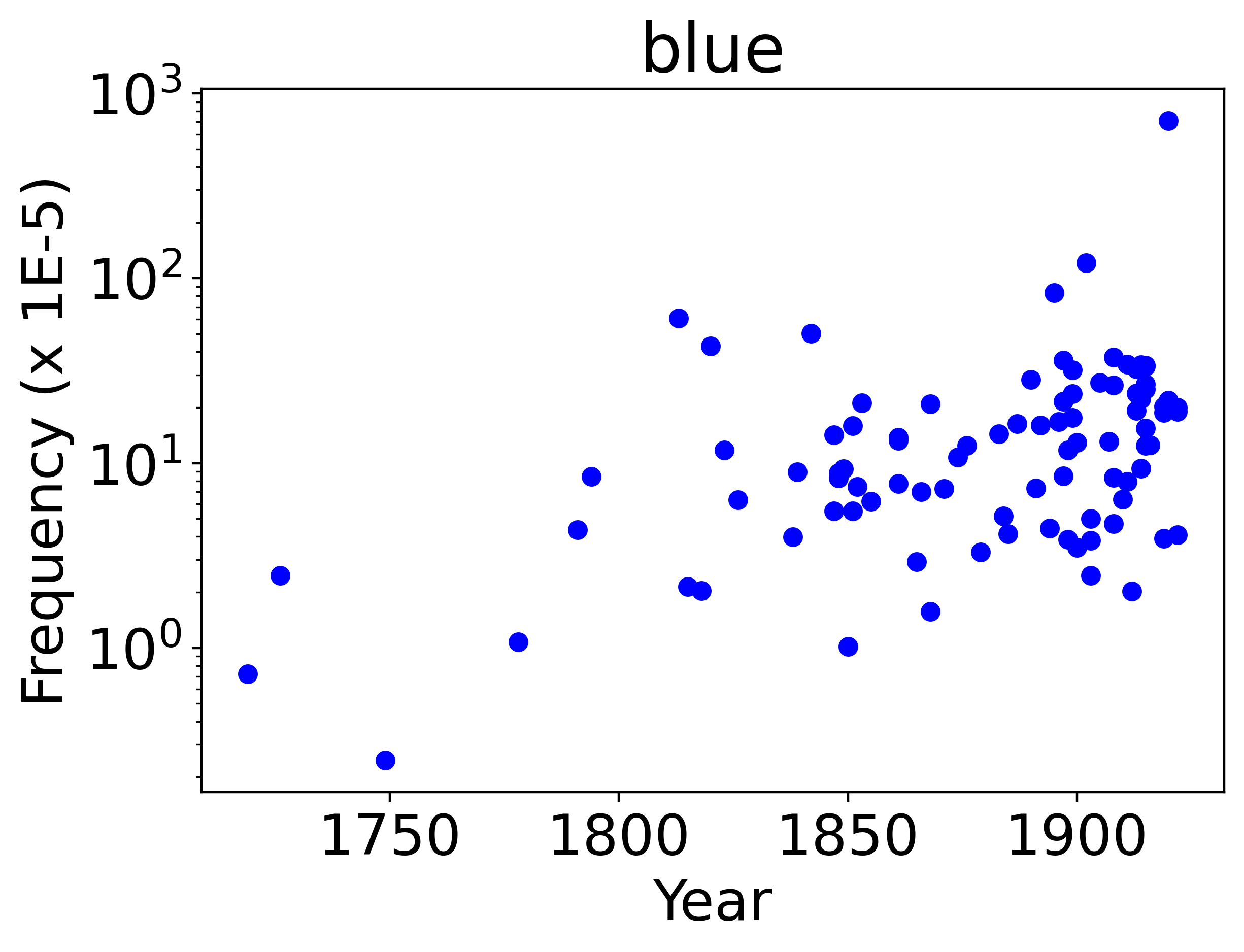}
    \includegraphics[scale=0.4]{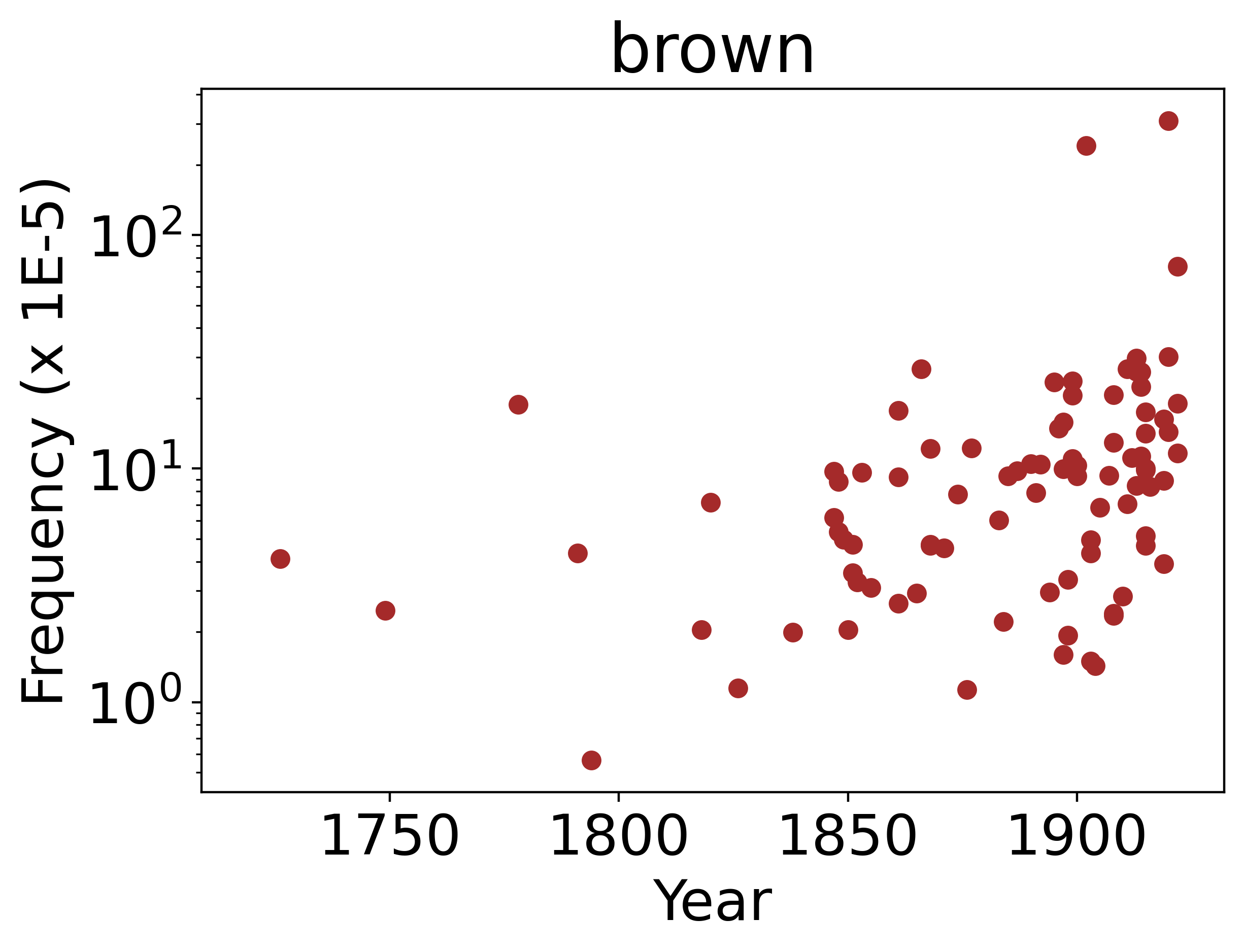}
    \includegraphics[scale=0.4]{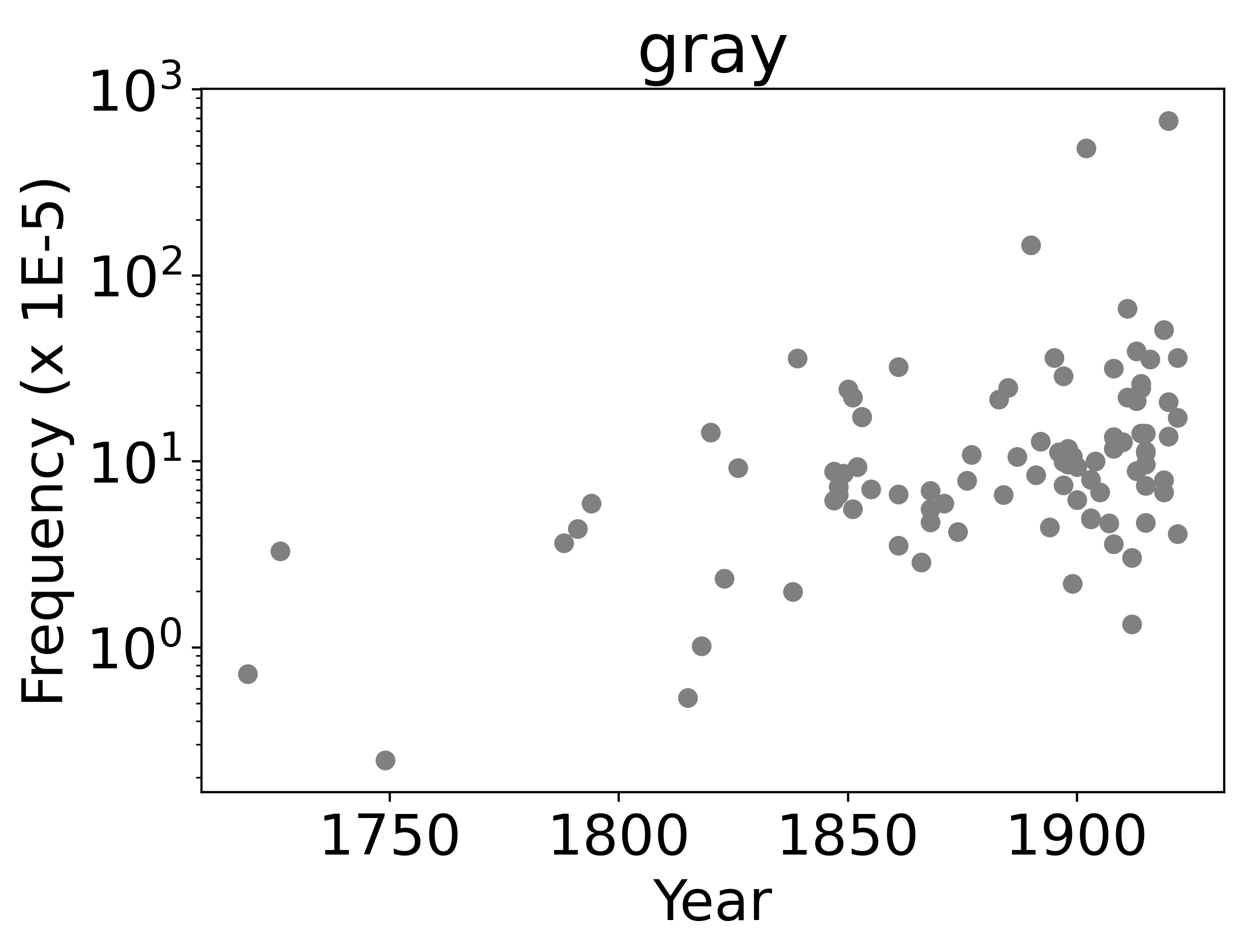}
    \includegraphics[scale=0.4]{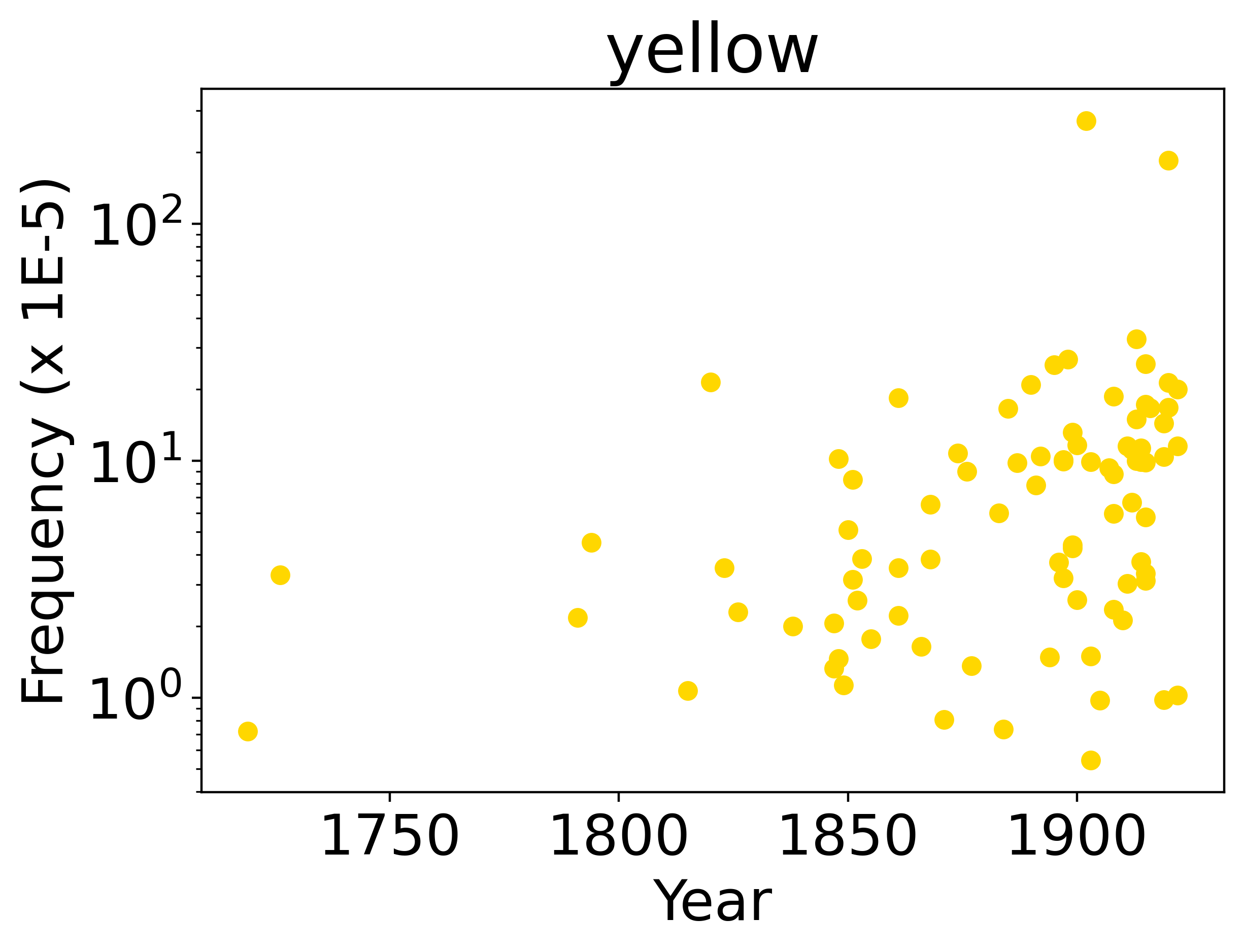}
    \includegraphics[scale=0.4]{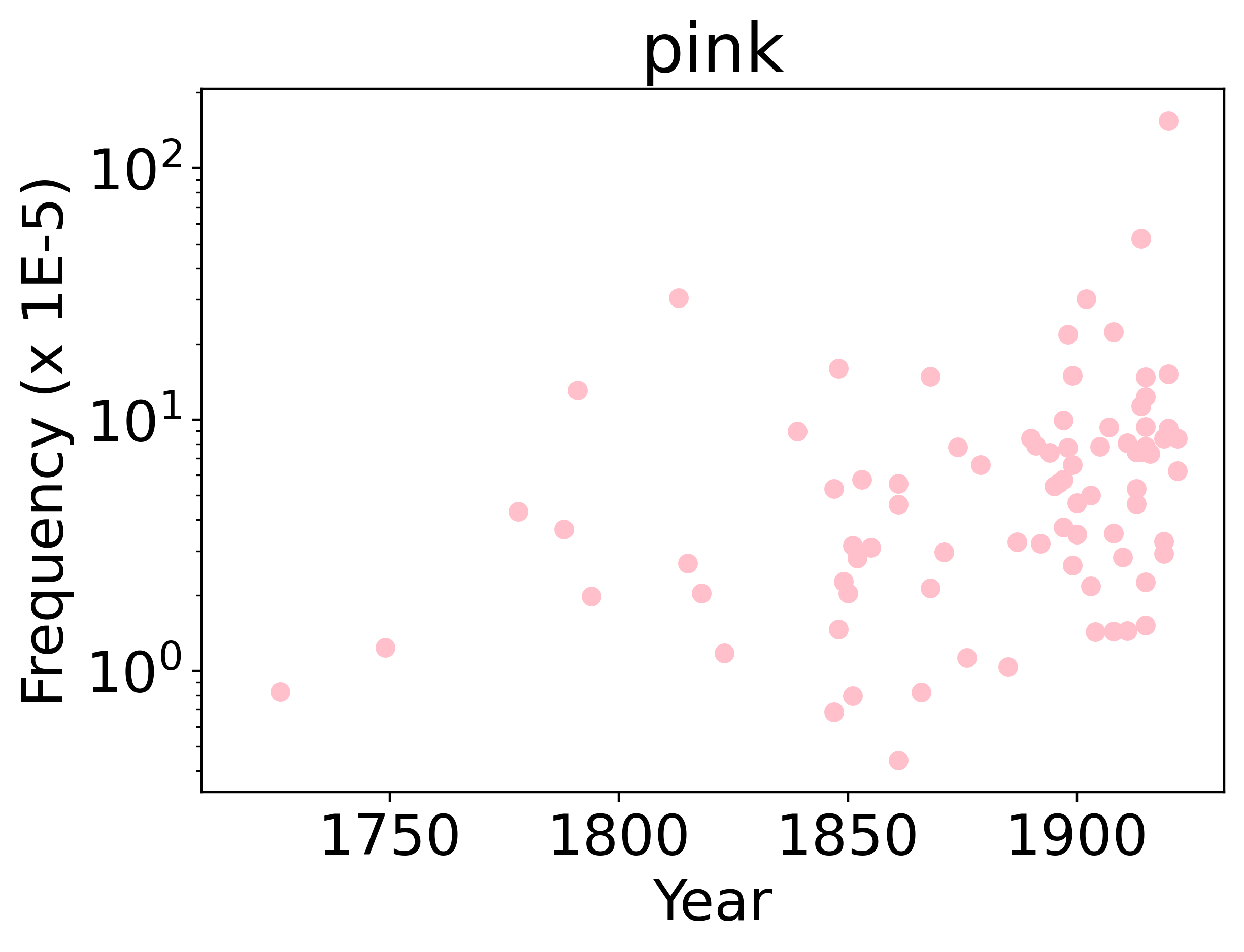}
    \includegraphics[scale=0.4]{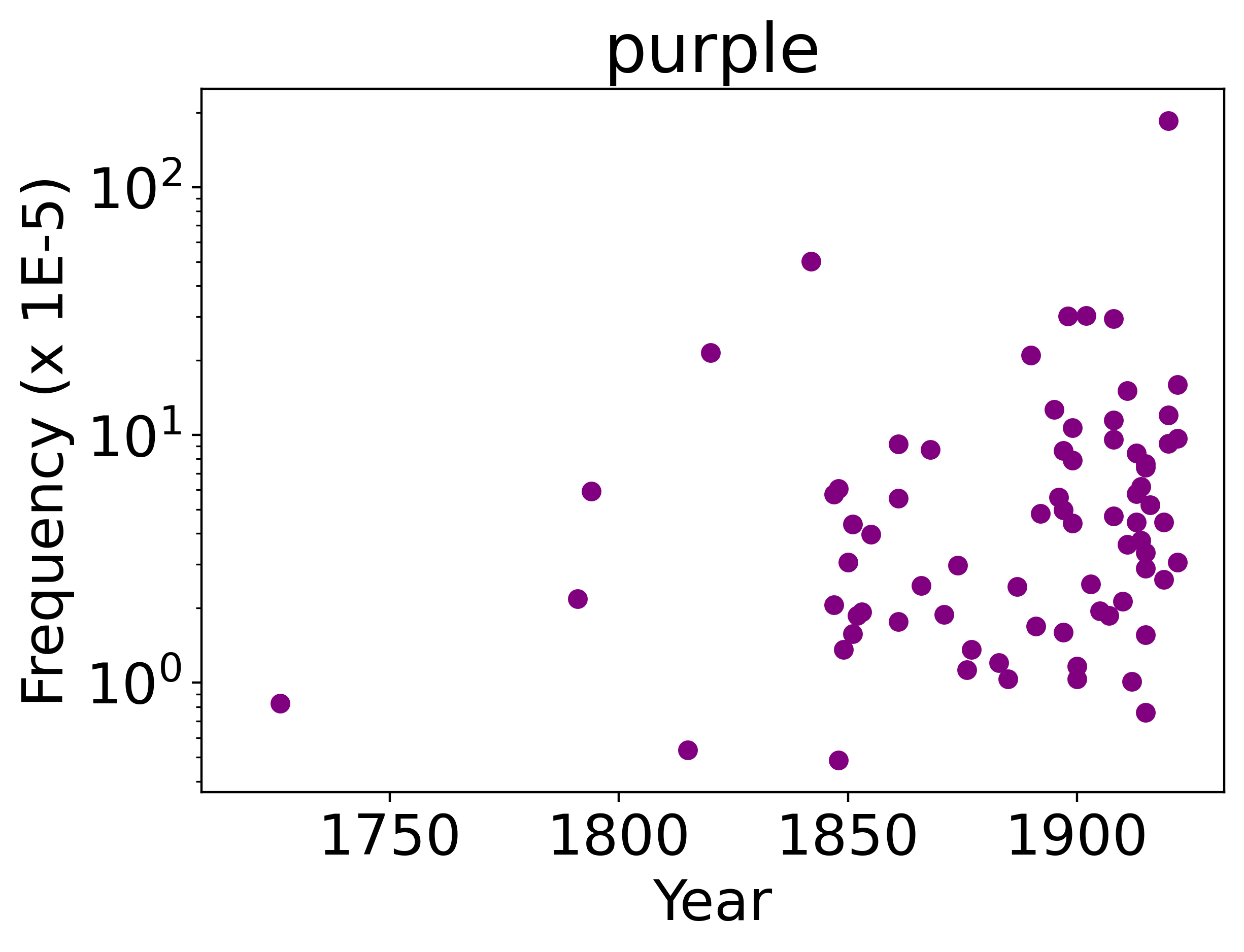}
    \caption{Log-scale scatter plots of normalized color term frequencies for each publication in LitBank.}
    \label{fig:color_freq}
\end{figure*}

We present the scatter plots of normalized frequencies in Figure~\ref{fig:color_freq}. 

\section{Glasgow Norm Plots of Modified Nouns Over Time} \label{app:norms}
\begin{figure*}
    \centering
    \includegraphics[scale=0.4]{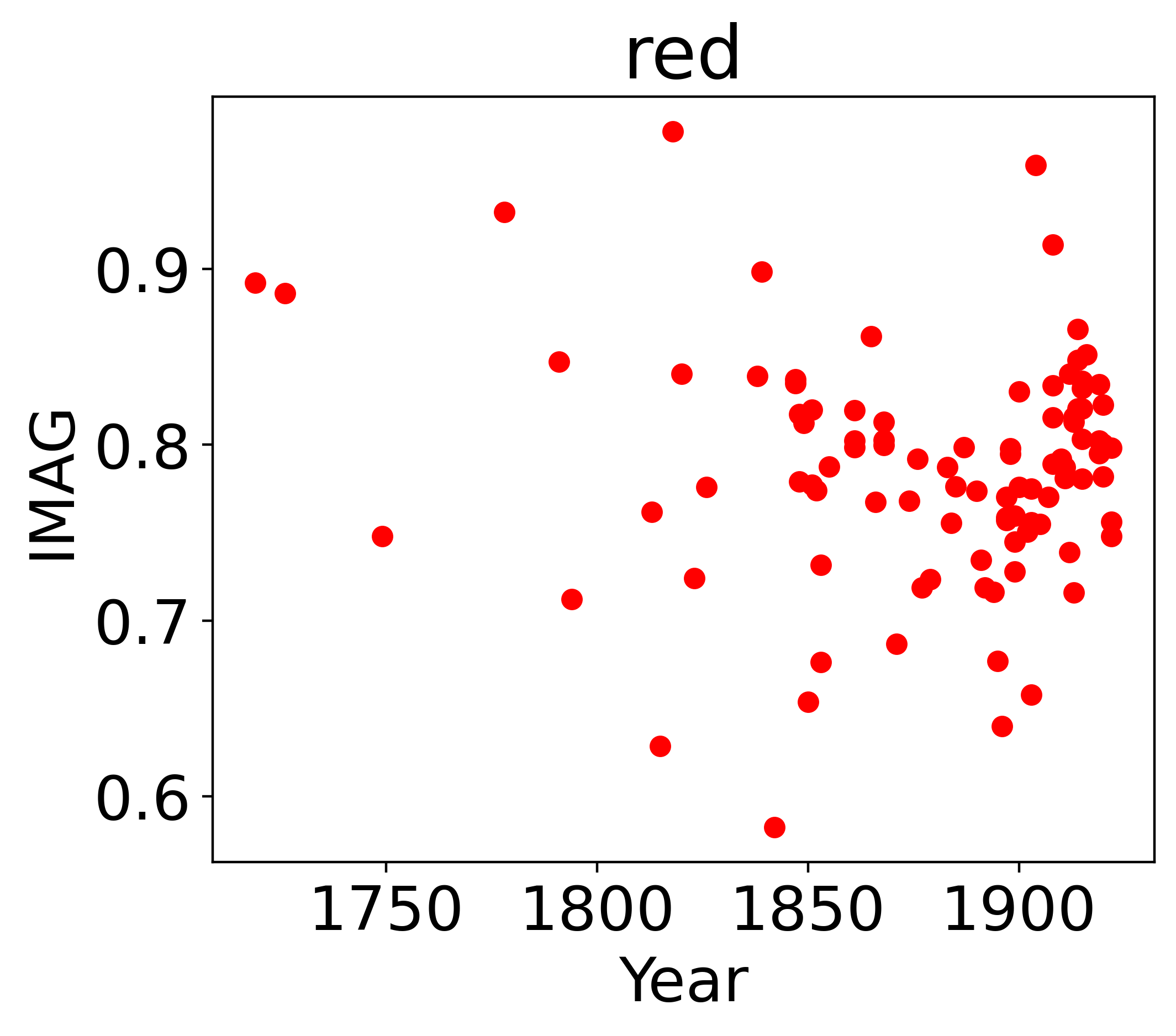}
    \includegraphics[scale=0.4]{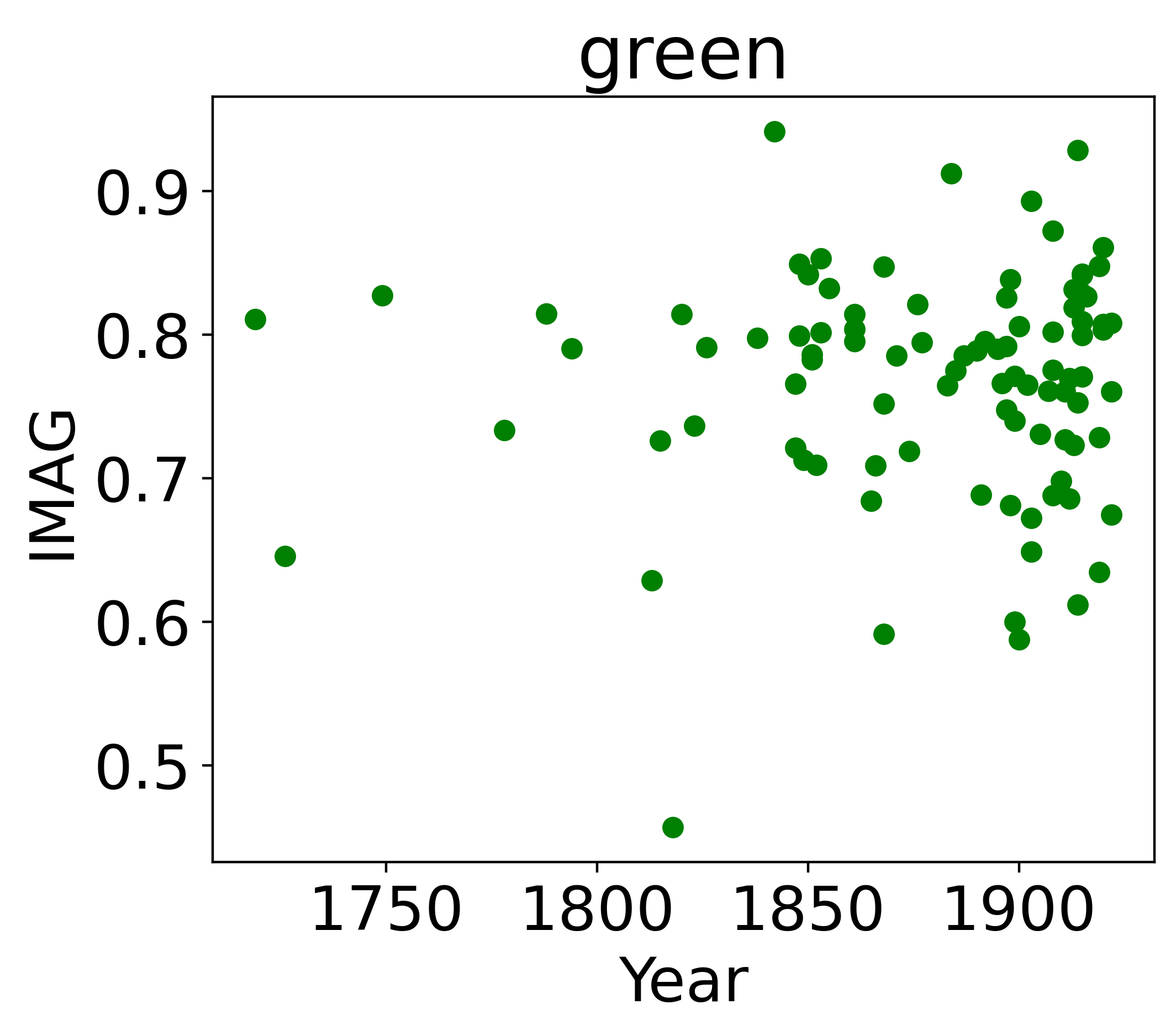}
    \includegraphics[scale=0.4]{figures/IMAG_black.png}
    \includegraphics[scale=0.4]{figures/IMAG_white.png}
    \includegraphics[scale=0.4]{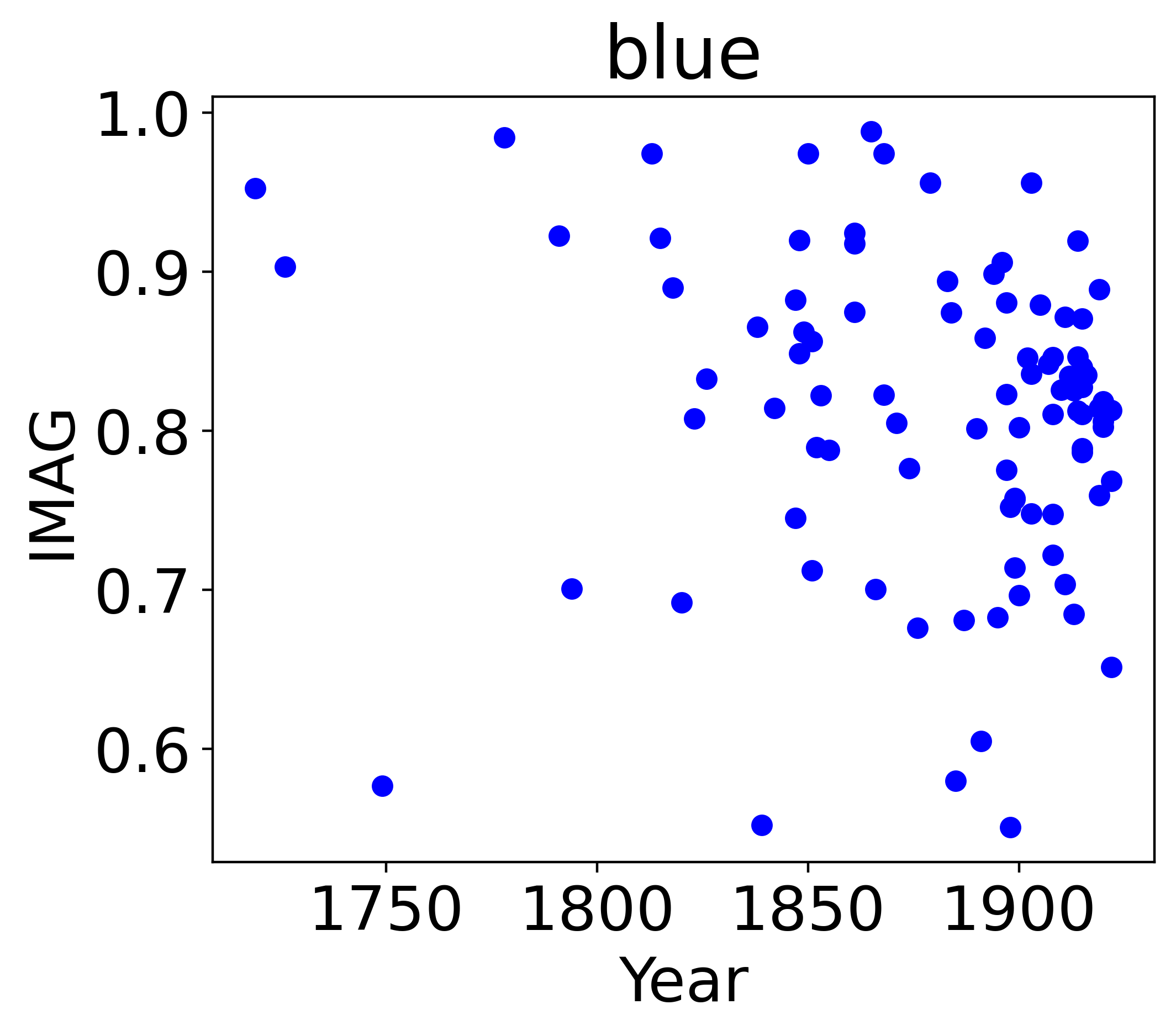}
    \includegraphics[scale=0.4]{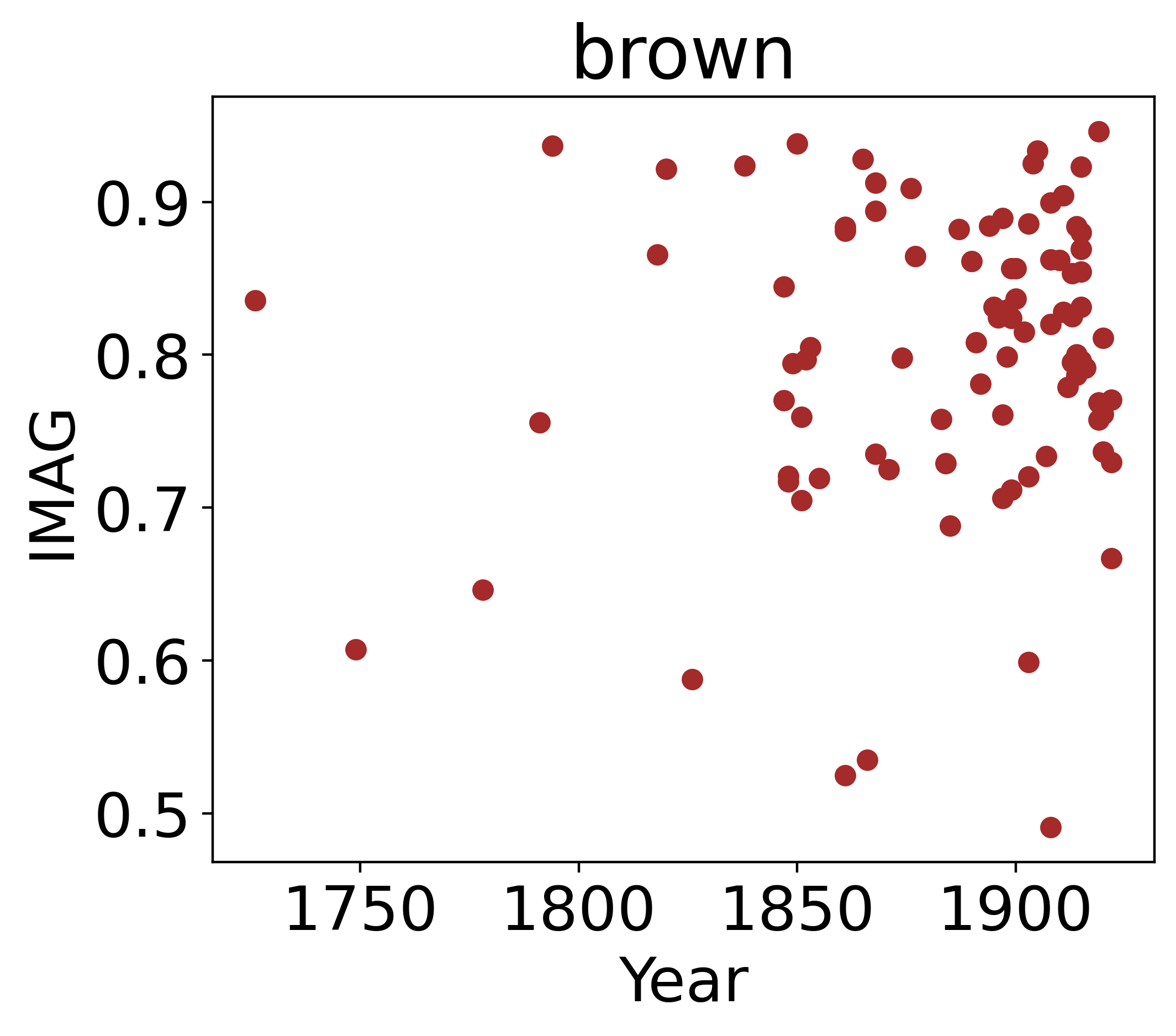}
    \includegraphics[scale=0.4]{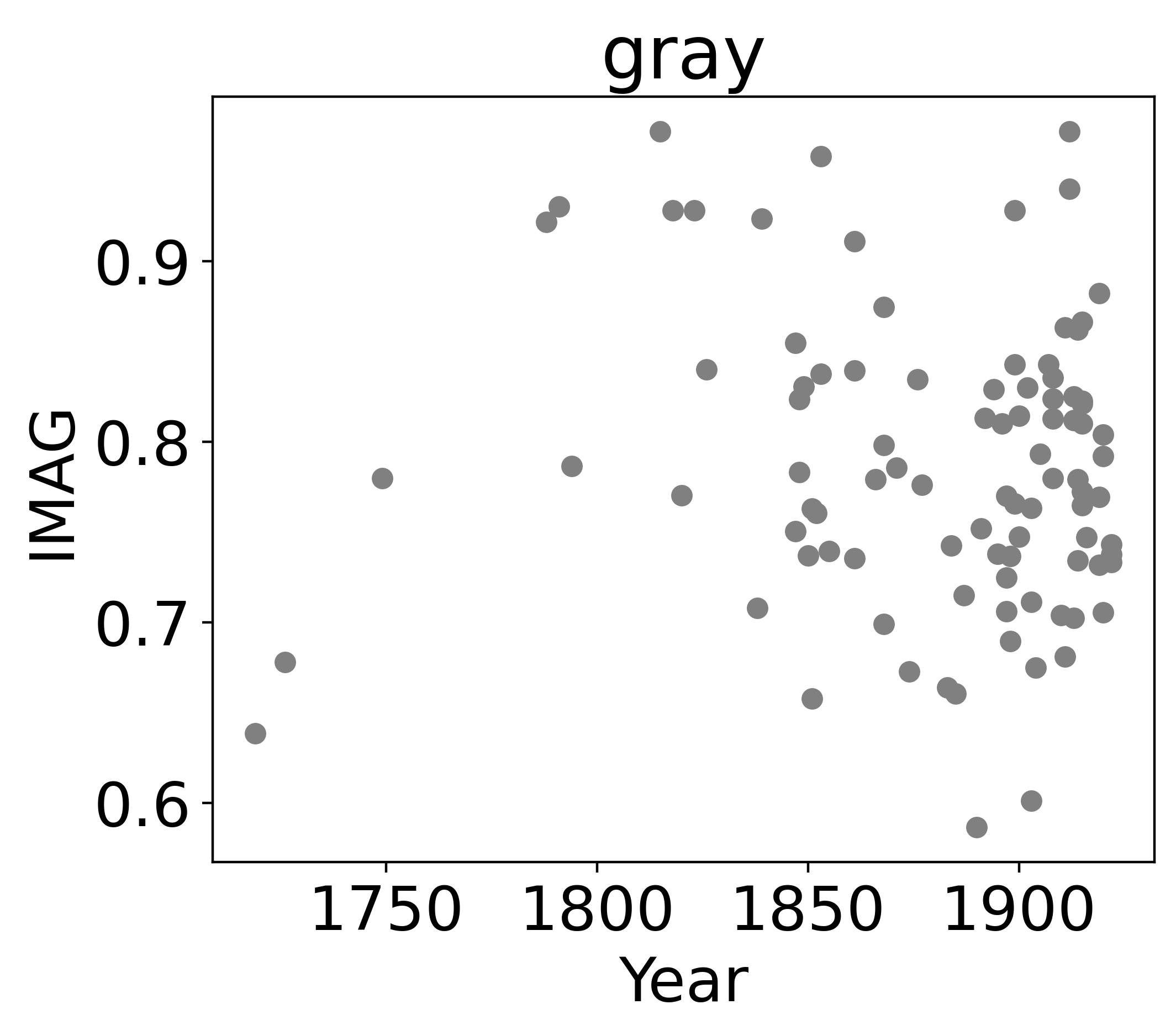}
    \includegraphics[scale=0.4]{figures/IMAG_yellow.png}
    \includegraphics[scale=0.4]{figures/IMAG_pink.png}
    \includegraphics[scale=0.4]{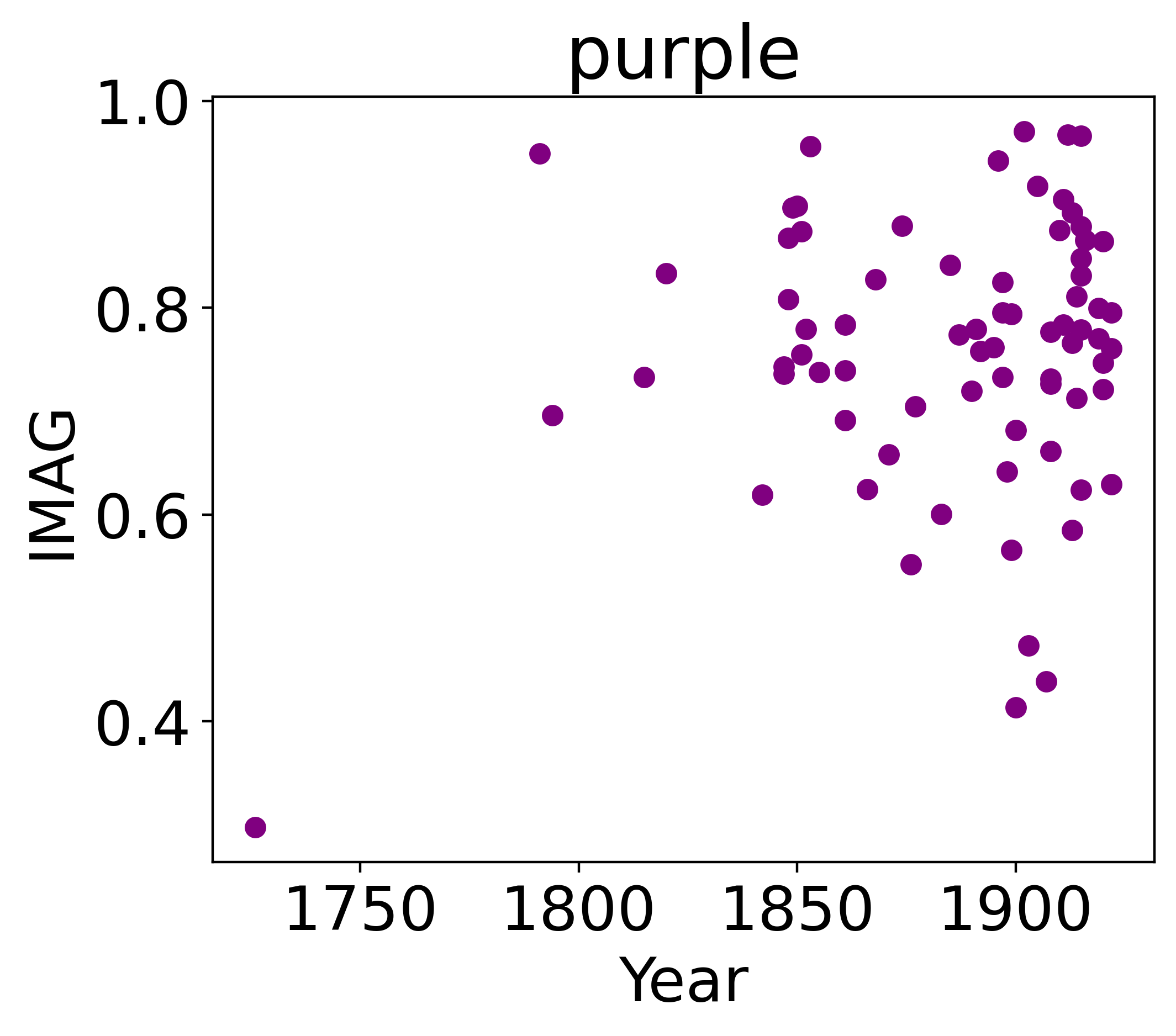}
    \caption{IMAG scatter plots of nouns modified by color terms for each publication in LitBank.}
    \label{fig:imag}
\end{figure*}

\begin{figure*}
    \centering
    \includegraphics[scale=0.4]{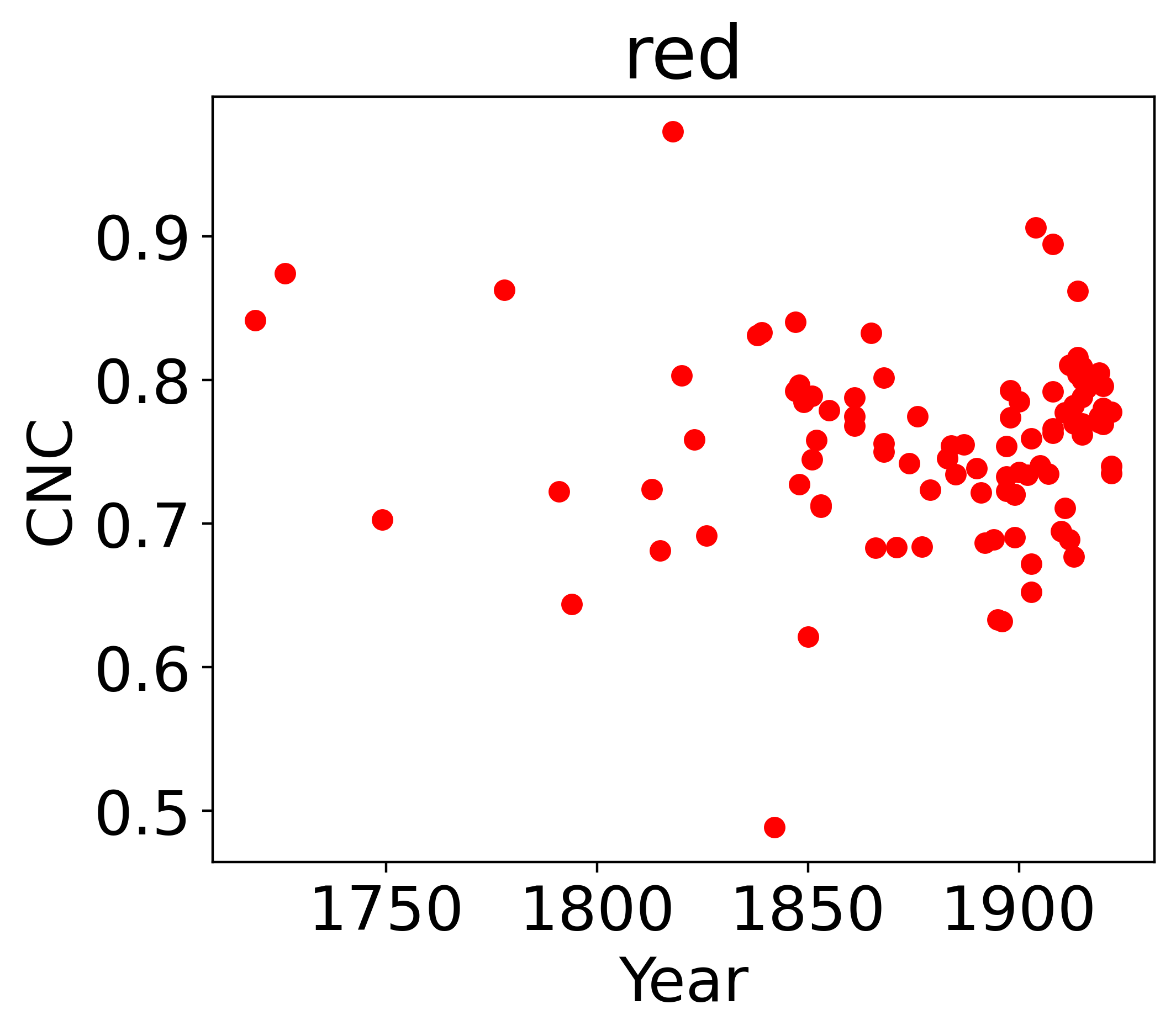}
    \includegraphics[scale=0.4]{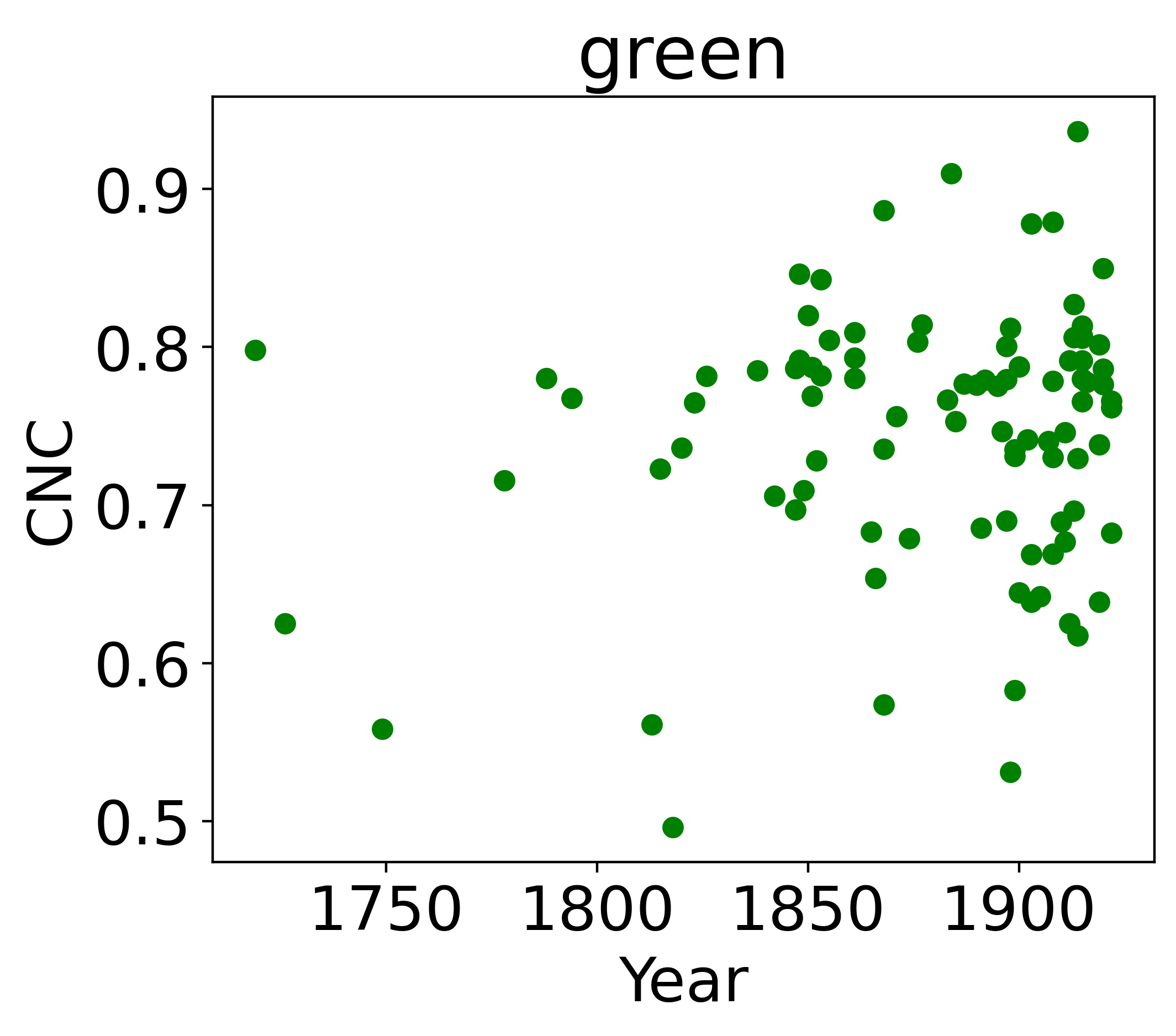}
    \includegraphics[scale=0.4]{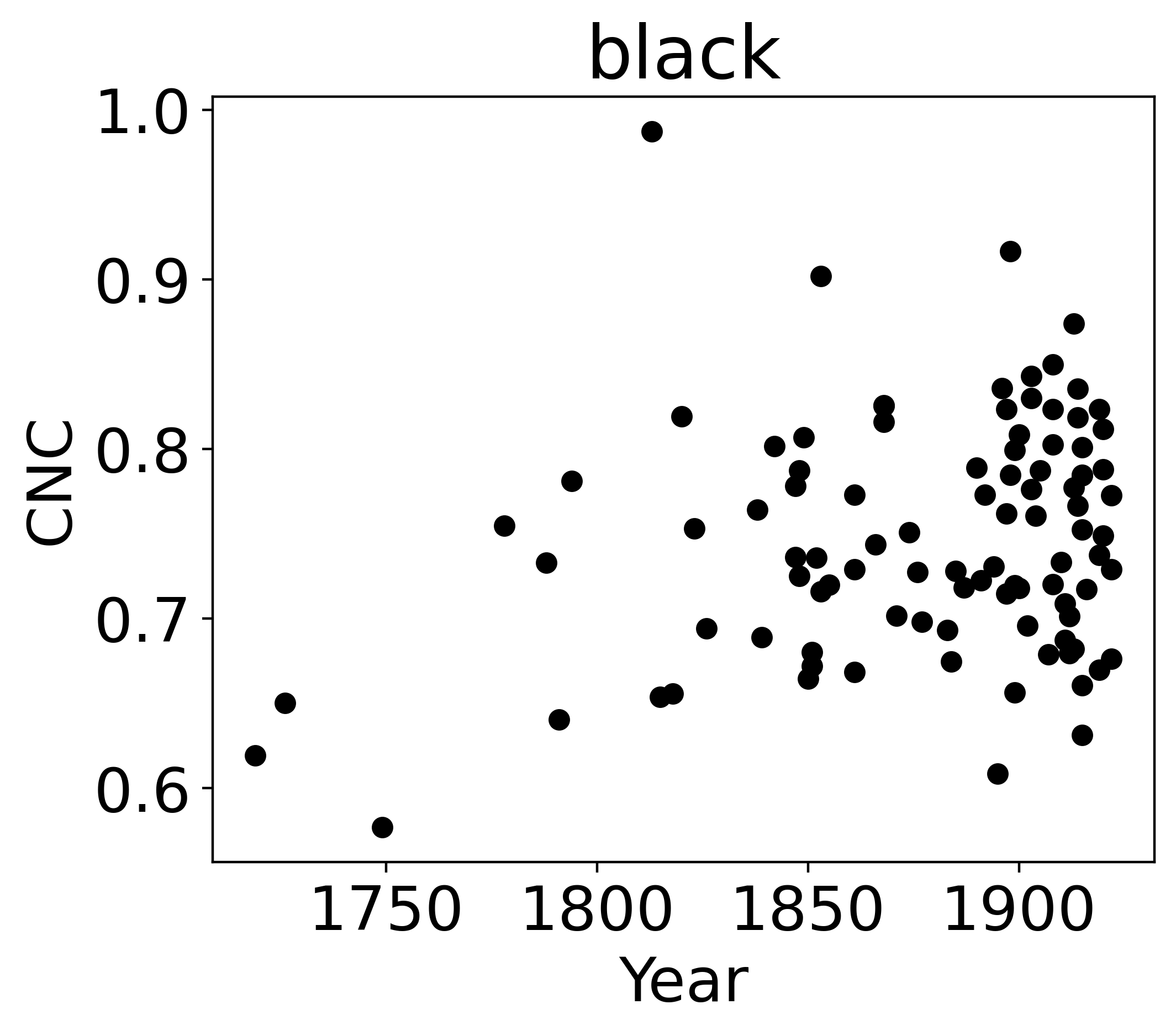}
    \includegraphics[scale=0.4]{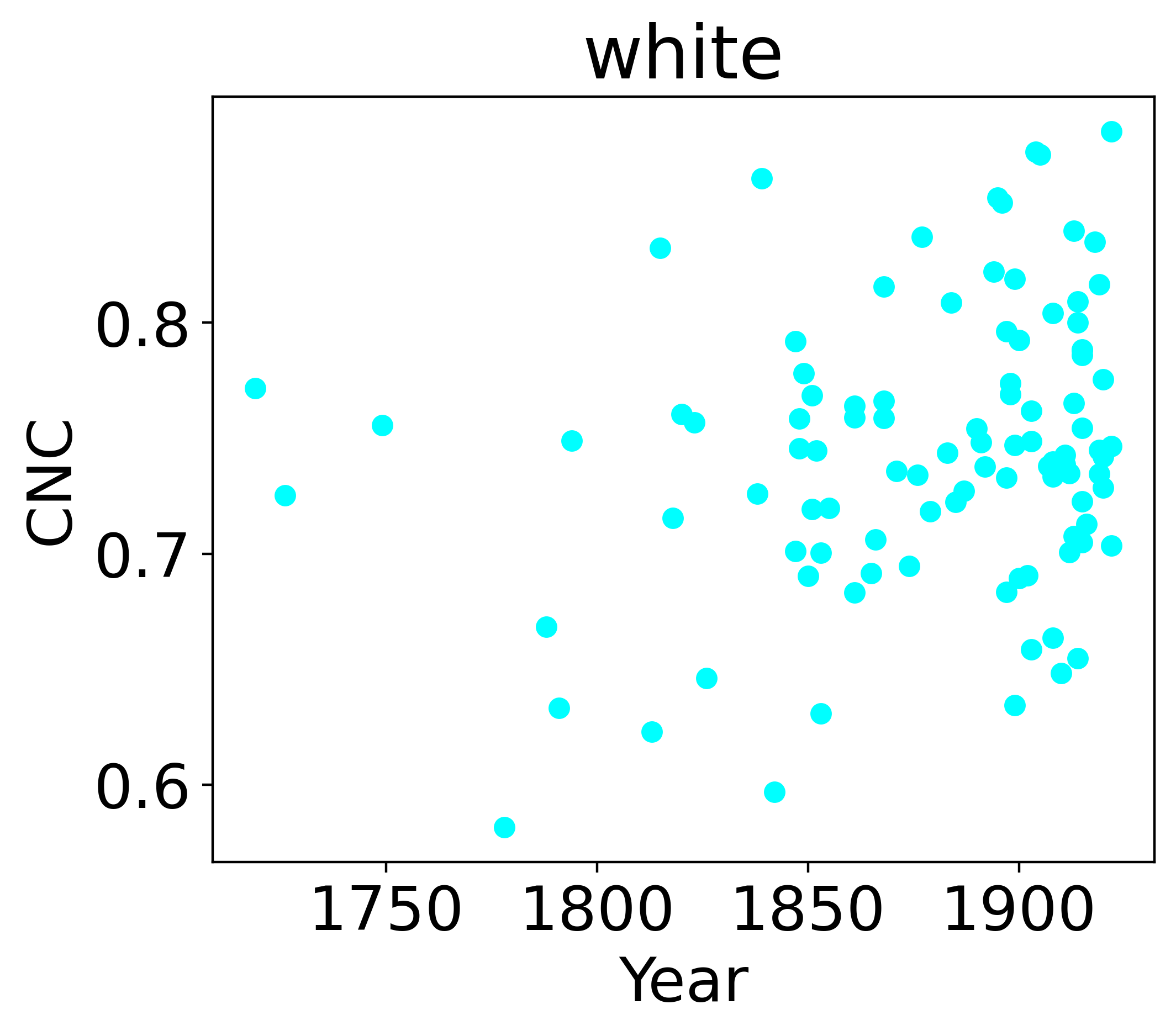}
    \includegraphics[scale=0.4]{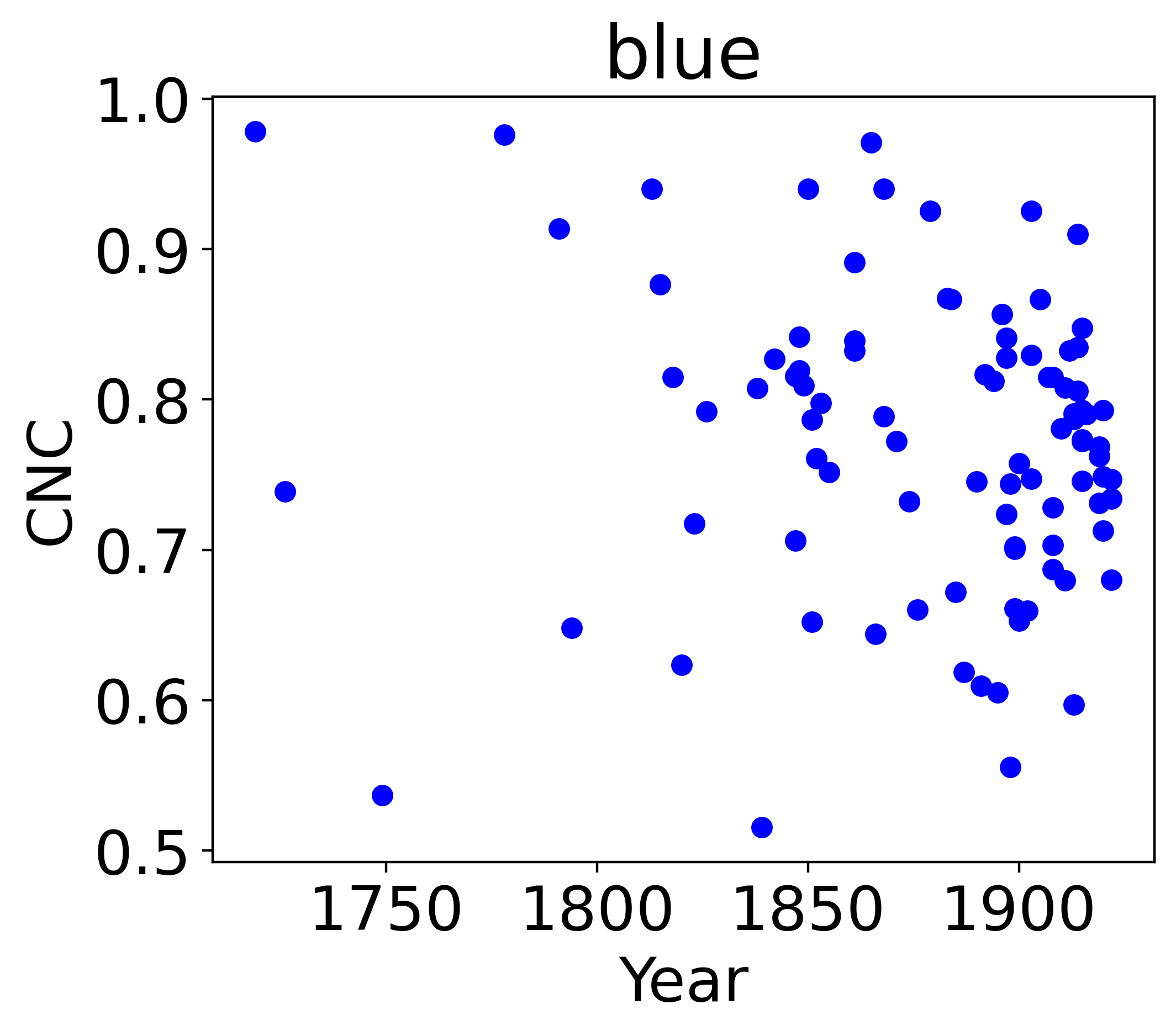}
    \includegraphics[scale=0.4]{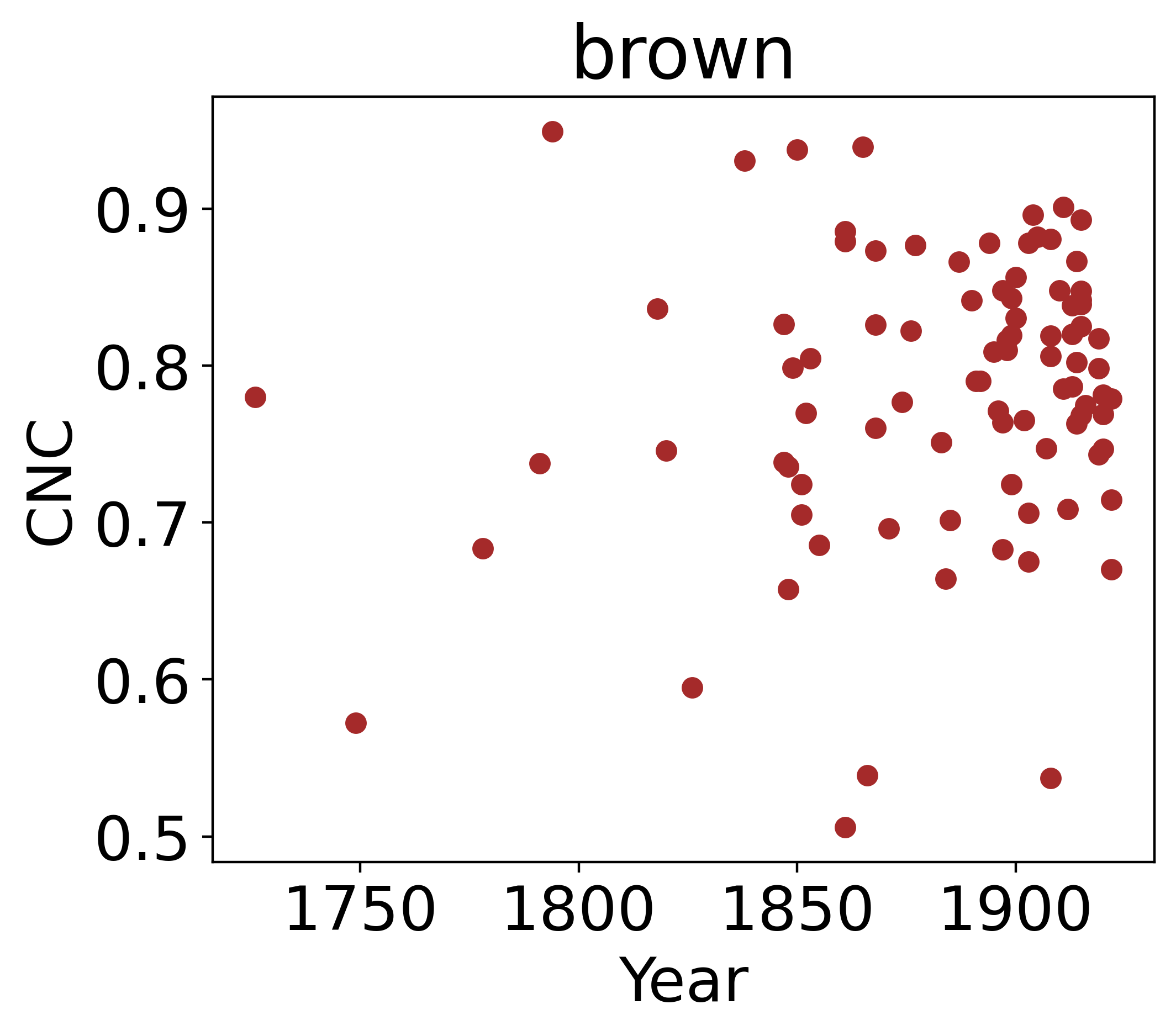}
    \includegraphics[scale=0.4]{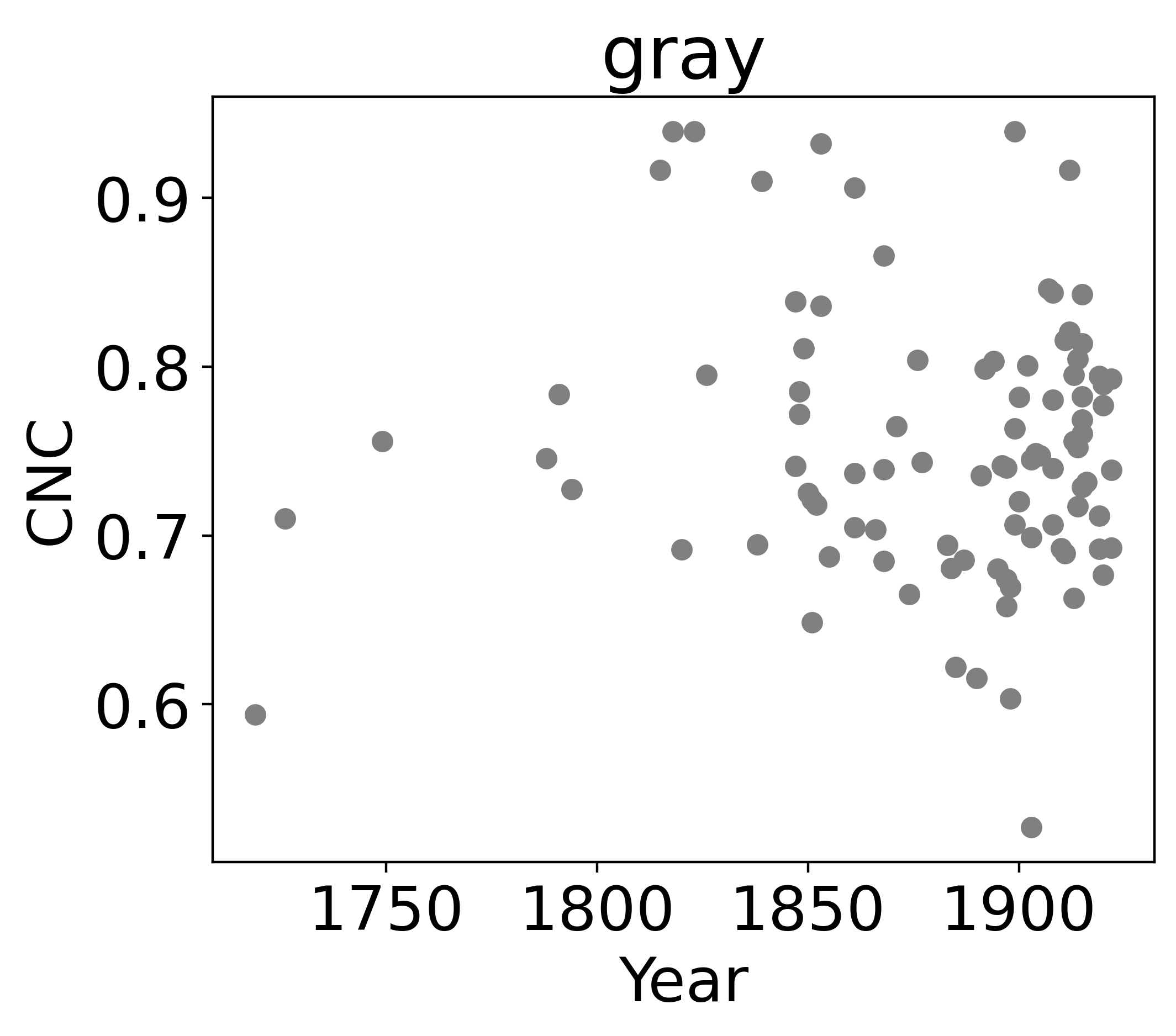}
    \includegraphics[scale=0.4]{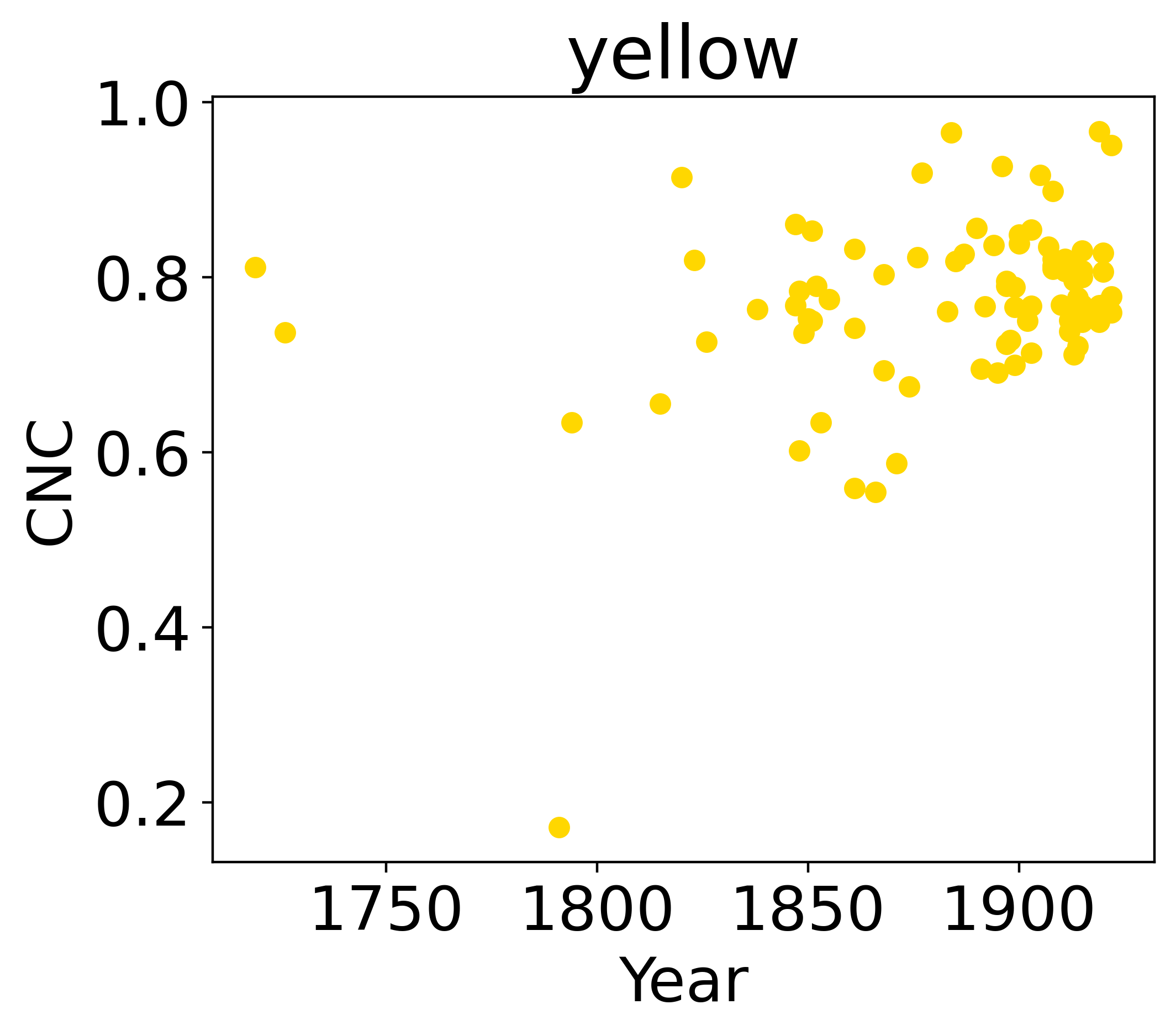}
    \includegraphics[scale=0.4]{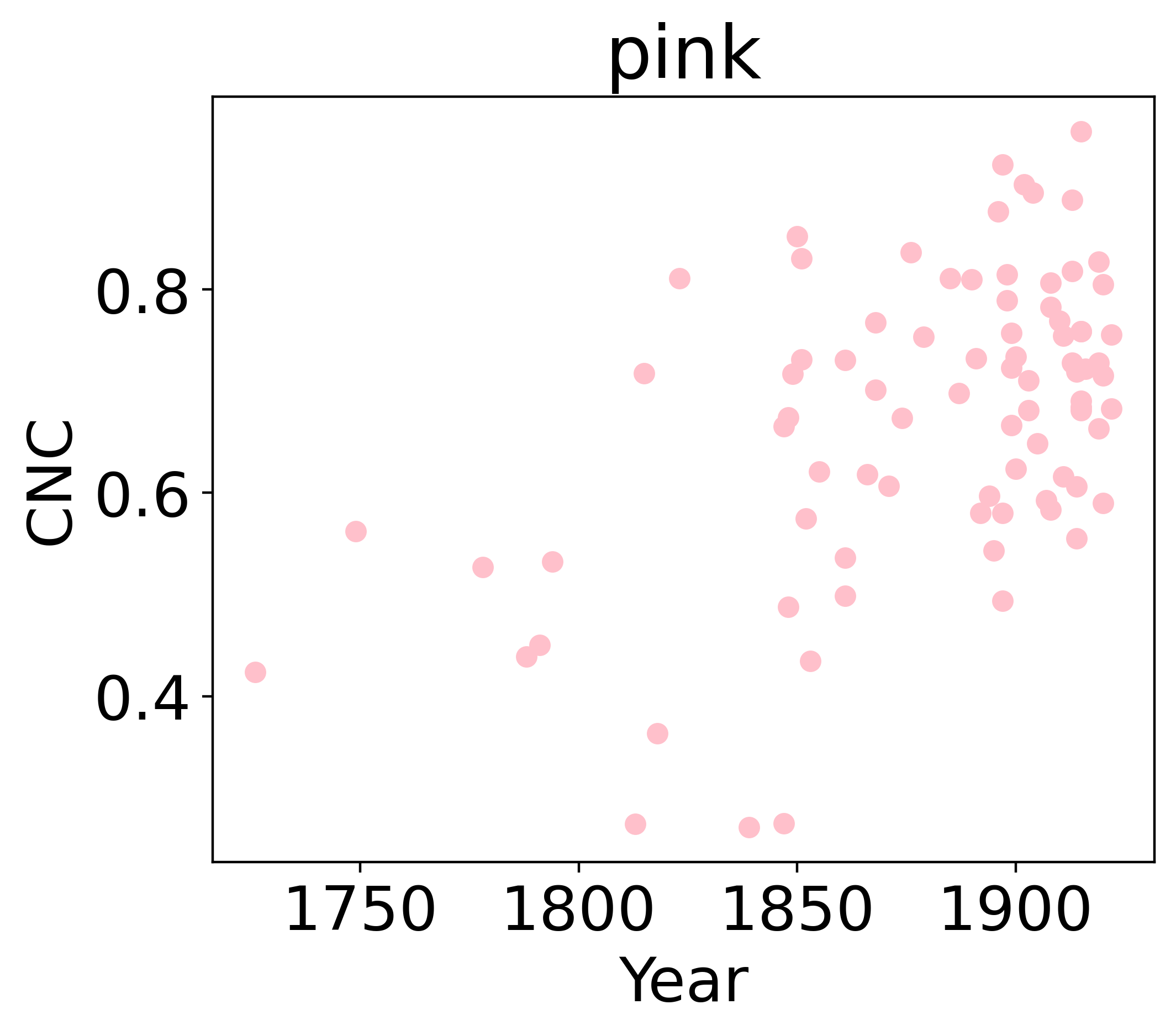}
    \includegraphics[scale=0.4]{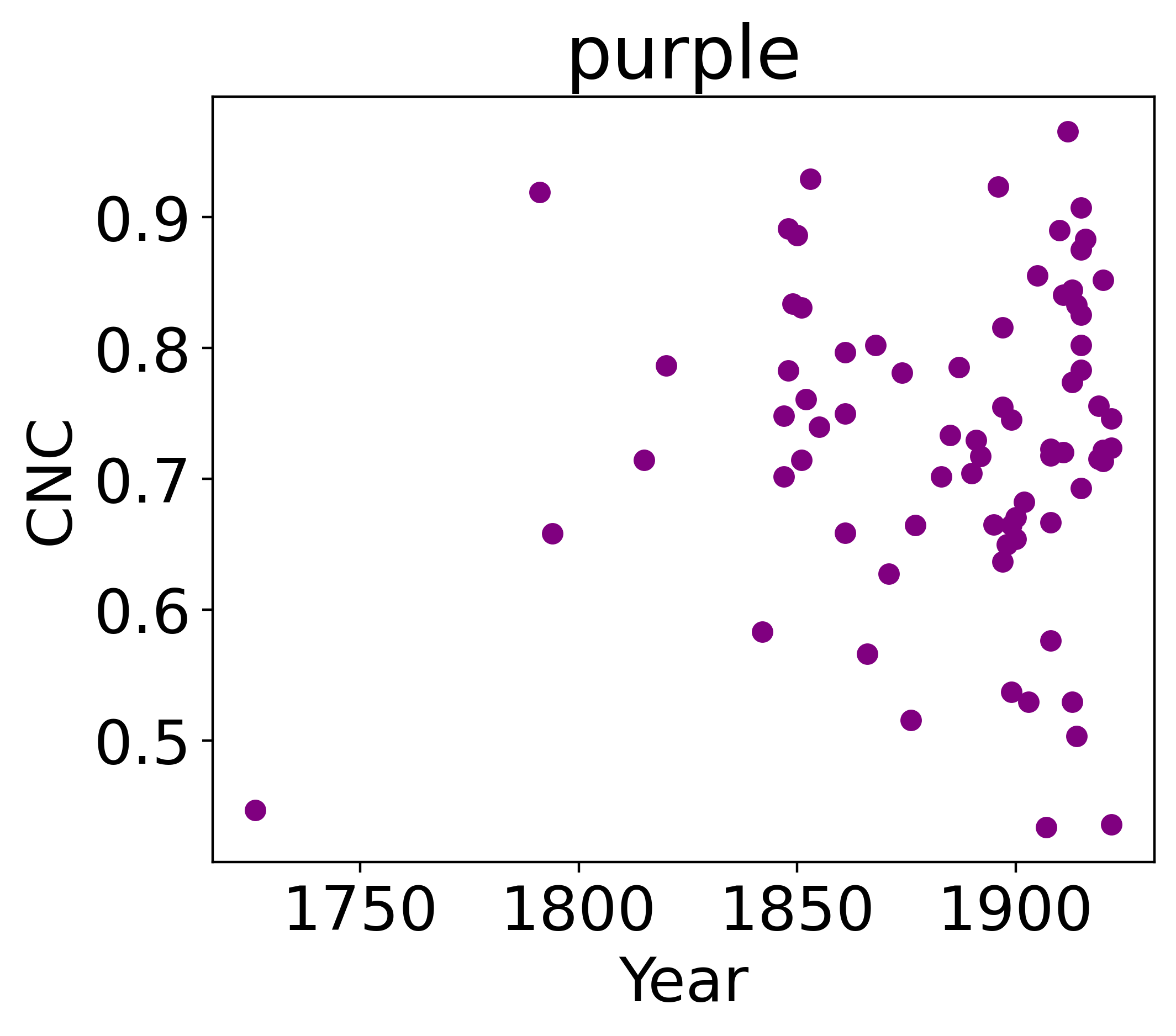}
    \caption{CNC scatter plots of nouns modified by color terms for each publication in LitBank.}
    \label{fig:cnc}
\end{figure*}

\begin{figure*}
    \centering
    \includegraphics[scale=0.4]{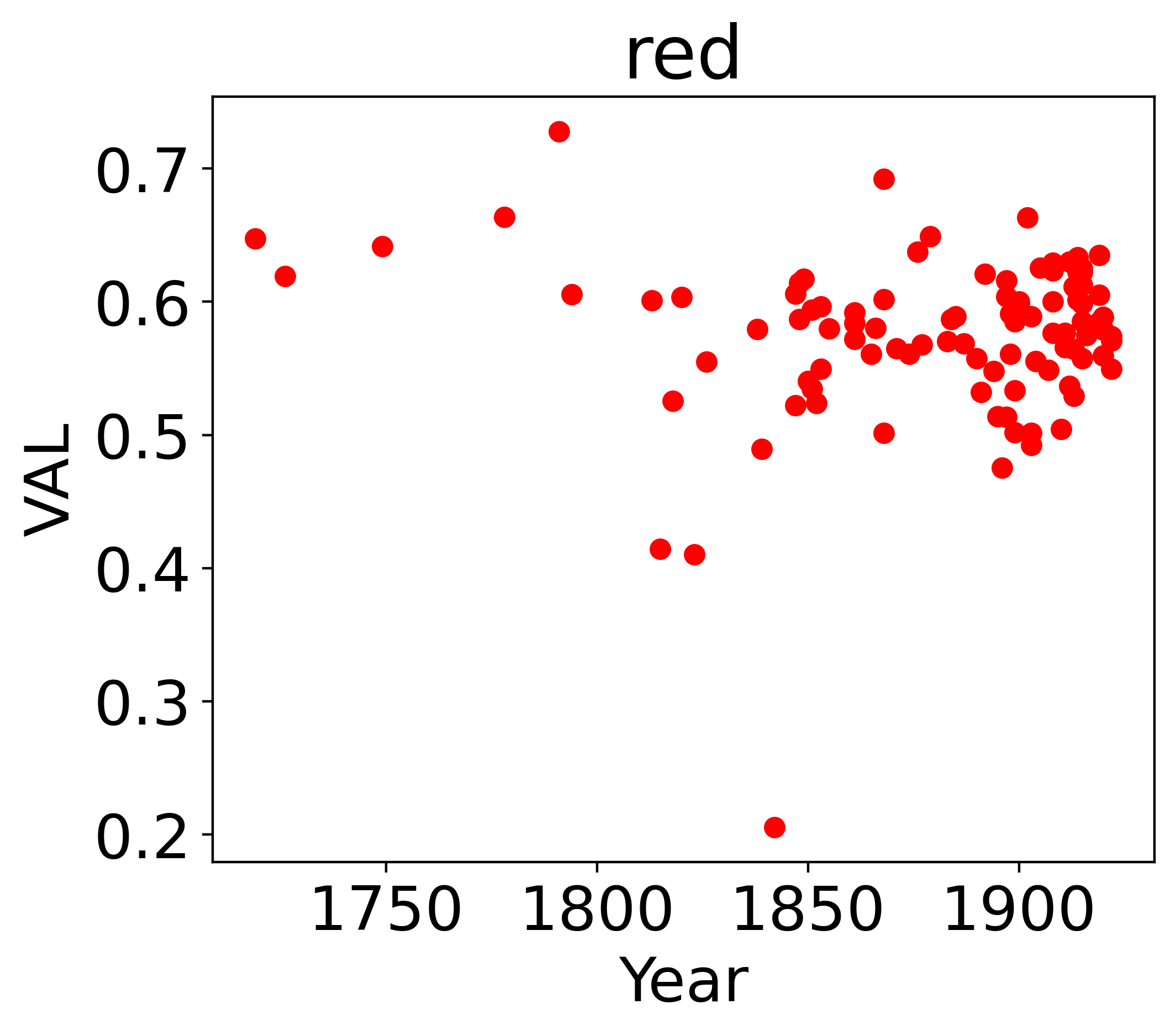}
    \includegraphics[scale=0.4]{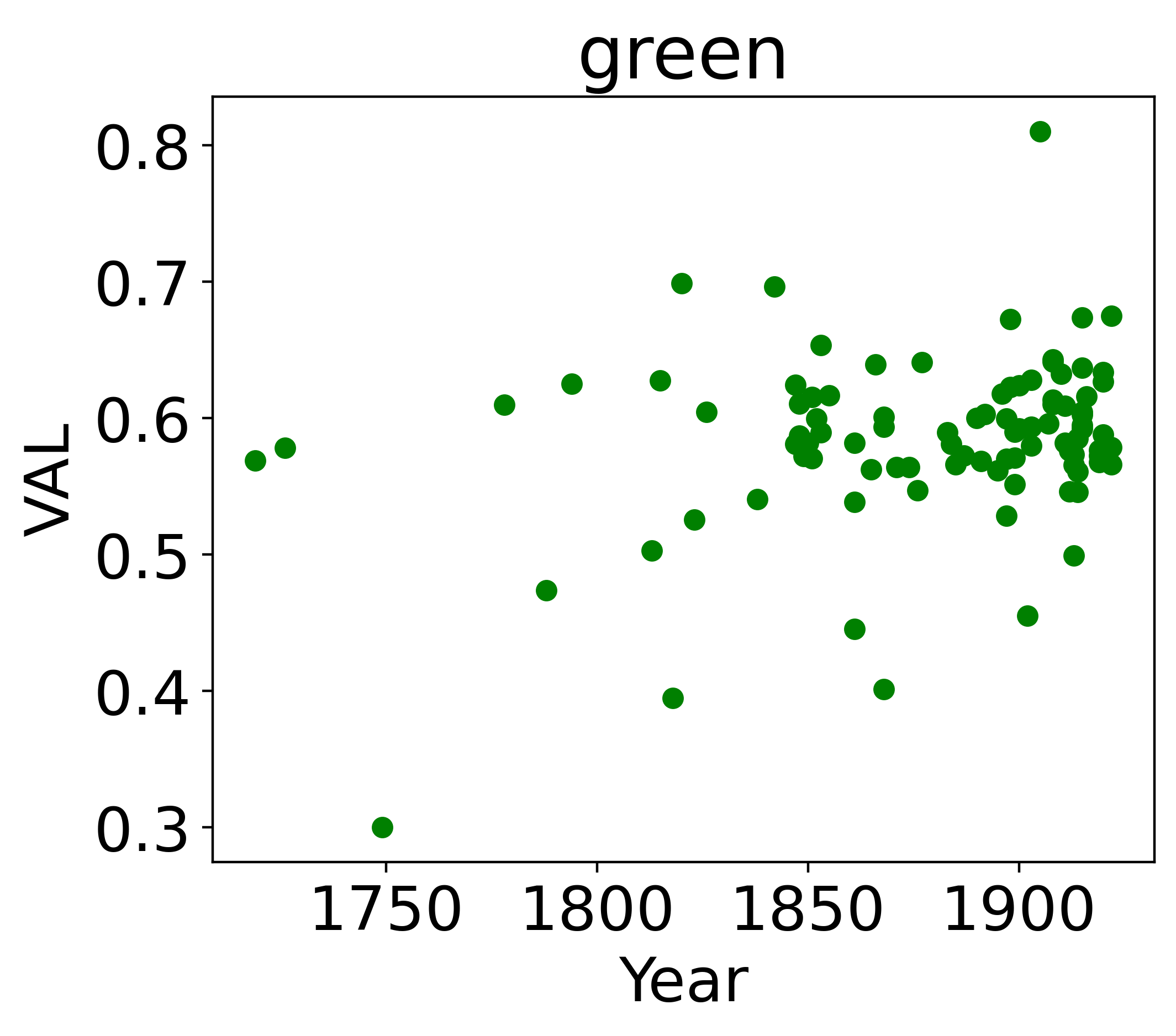}
    \includegraphics[scale=0.4]{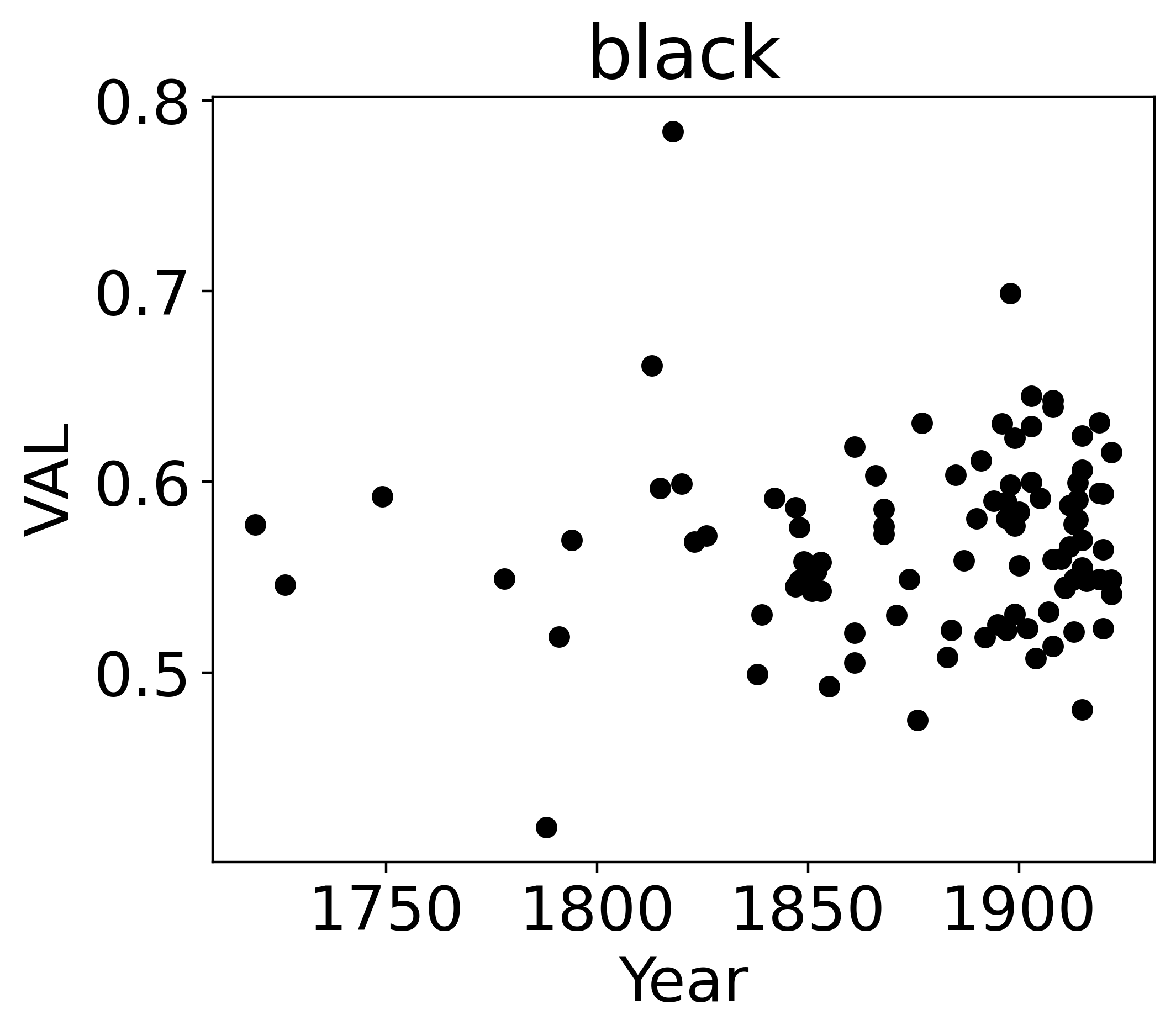}
    \includegraphics[scale=0.4]{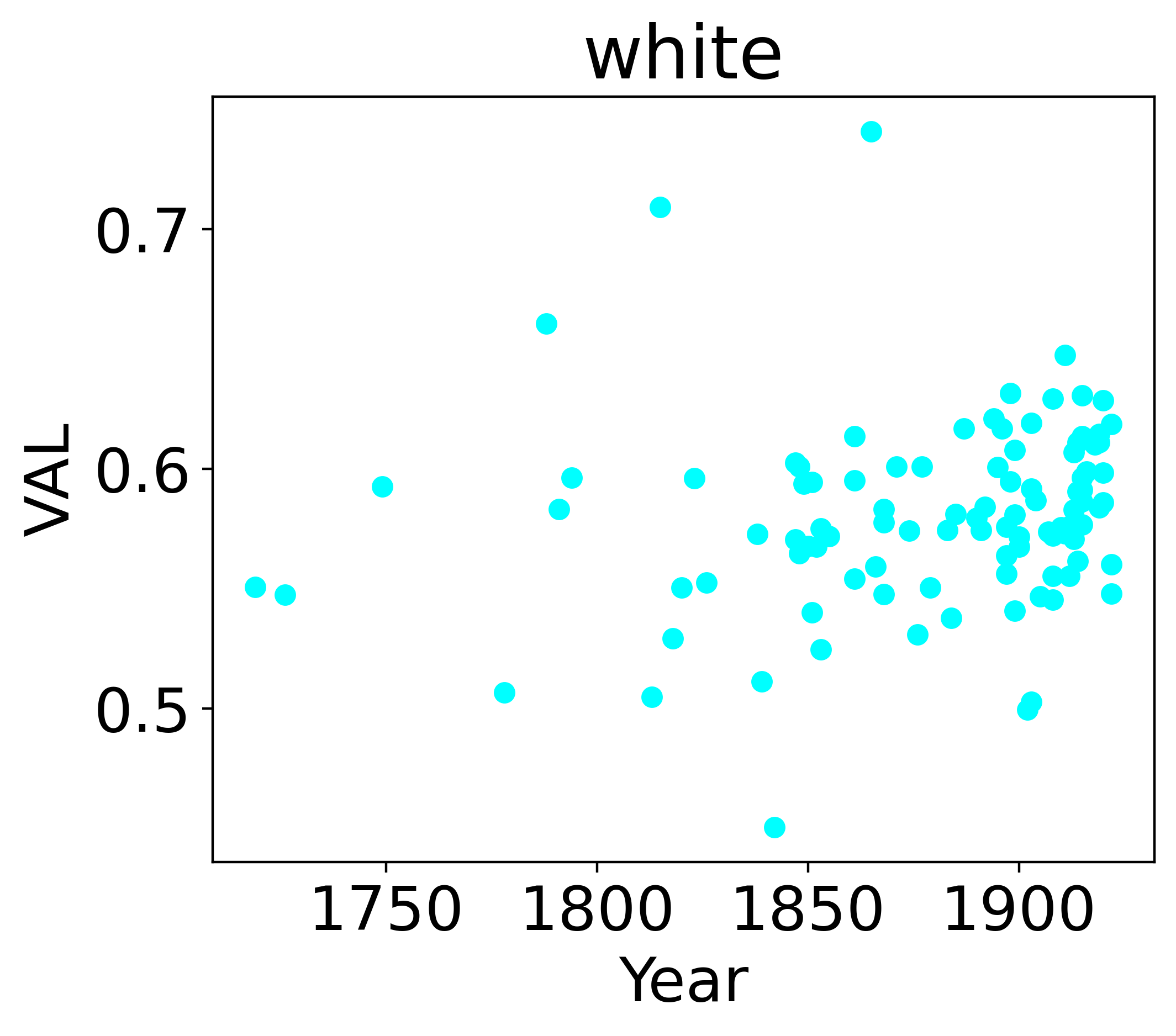}
    \includegraphics[scale=0.4]{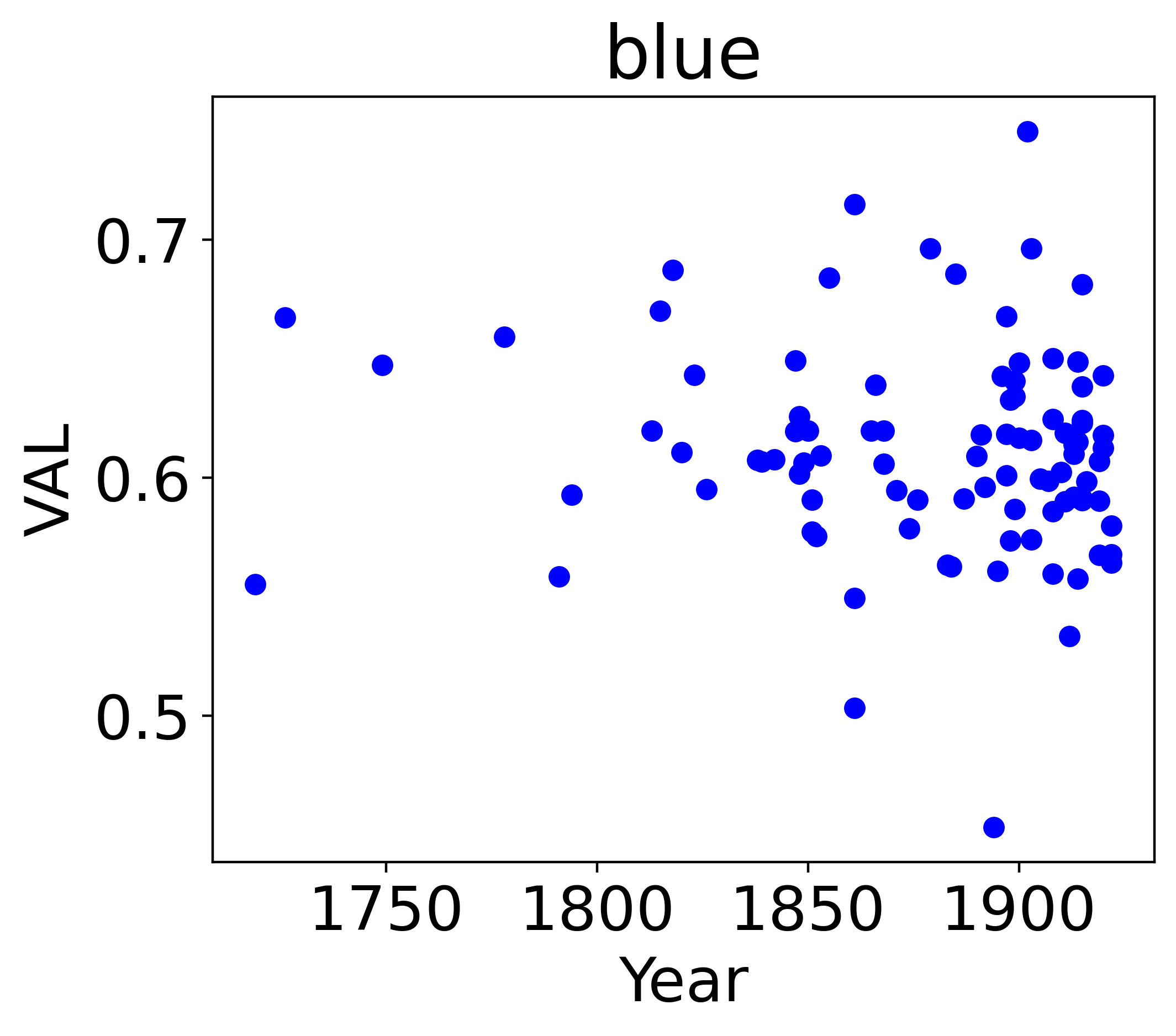}
    \includegraphics[scale=0.4]{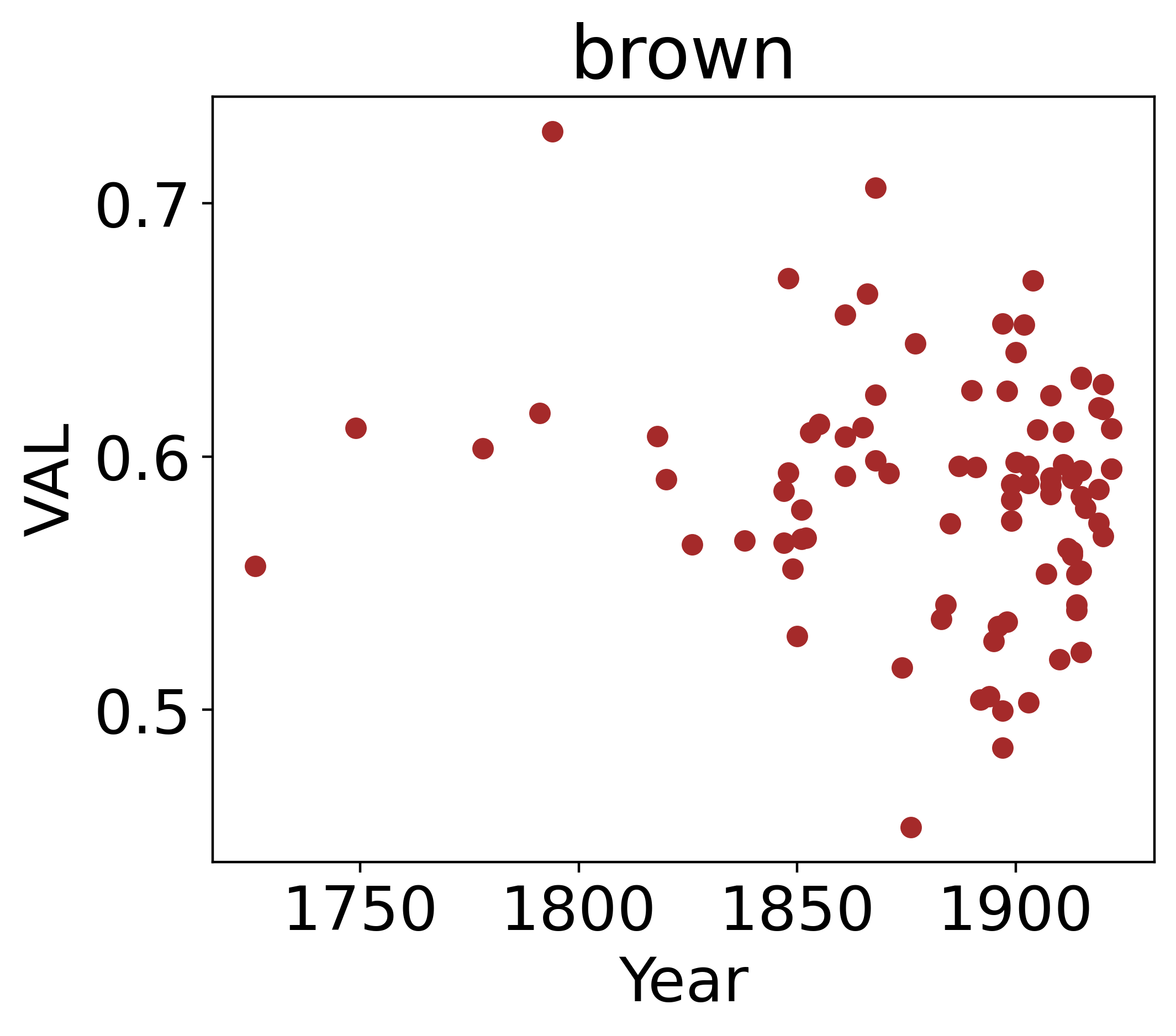}
    \includegraphics[scale=0.4]{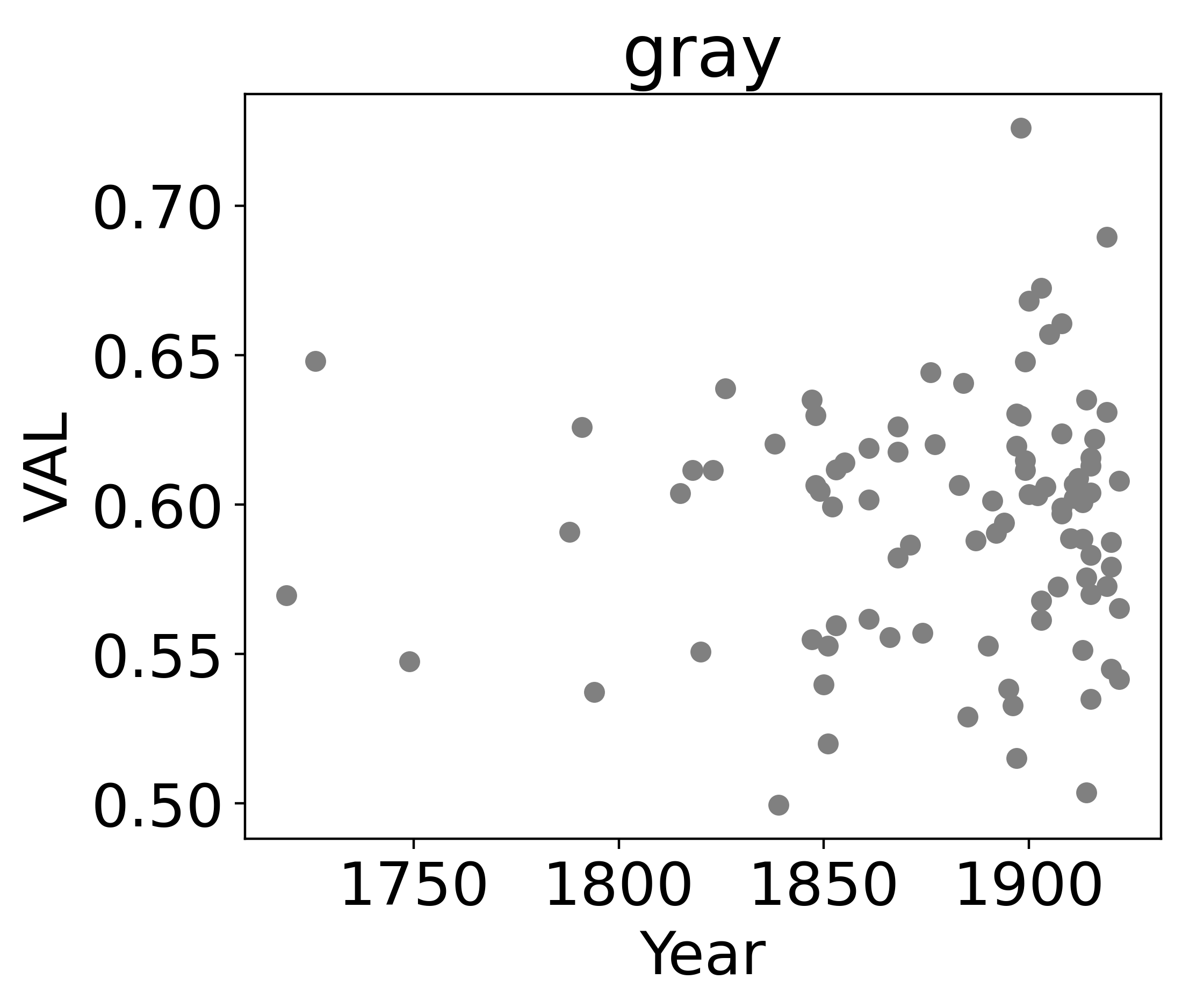}
    \includegraphics[scale=0.4]{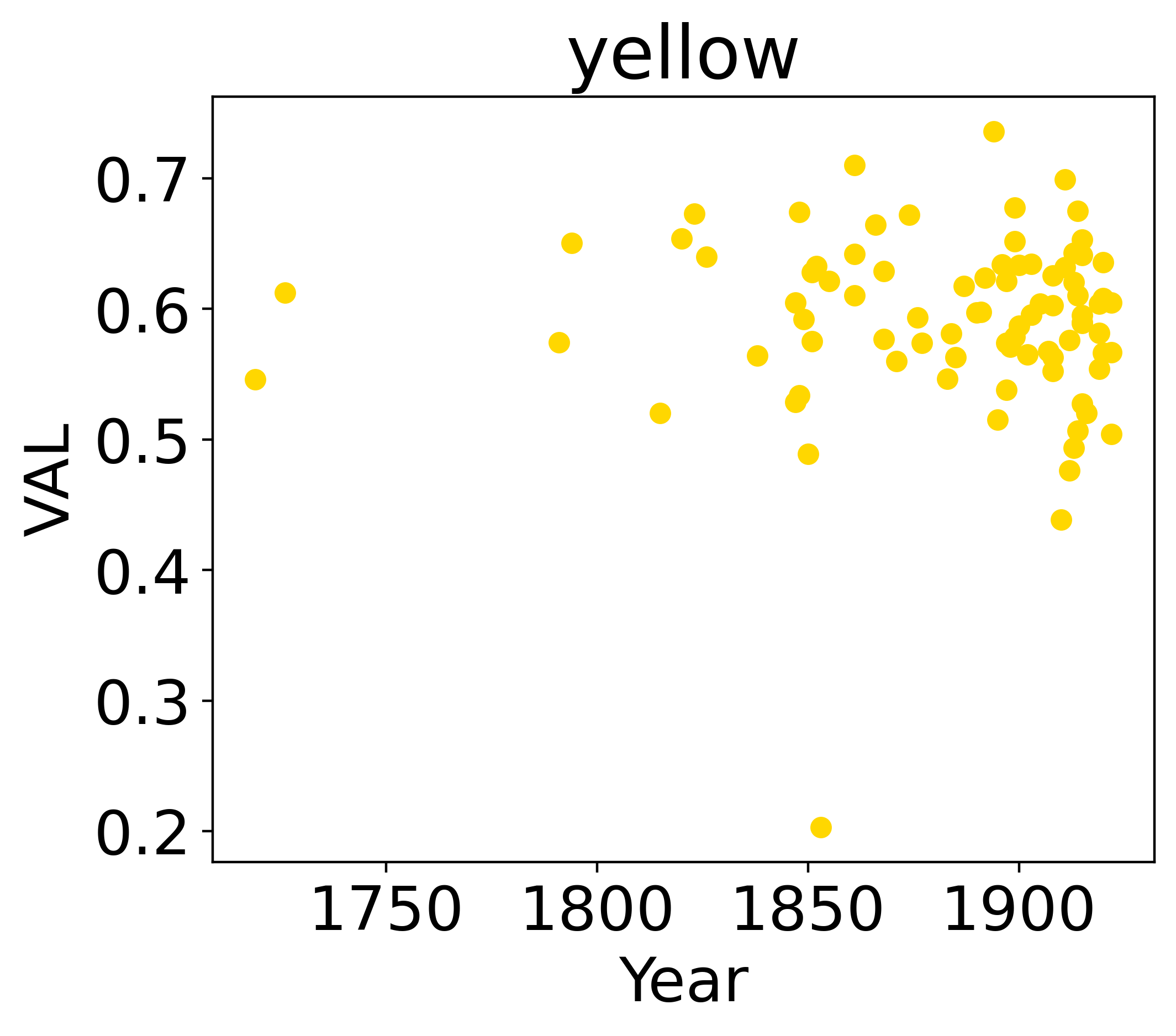}
    \includegraphics[scale=0.4]{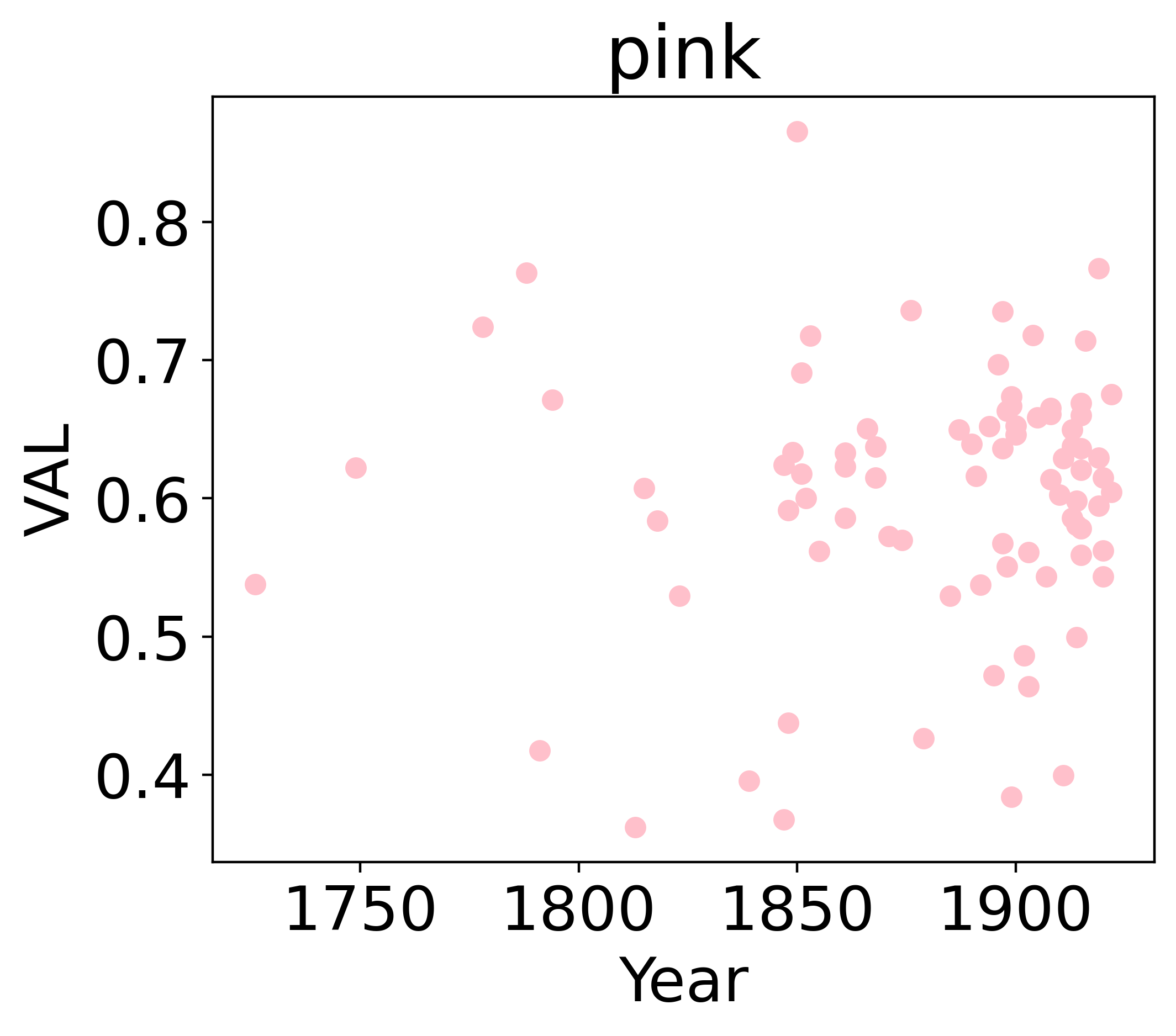}
    \includegraphics[scale=0.4]{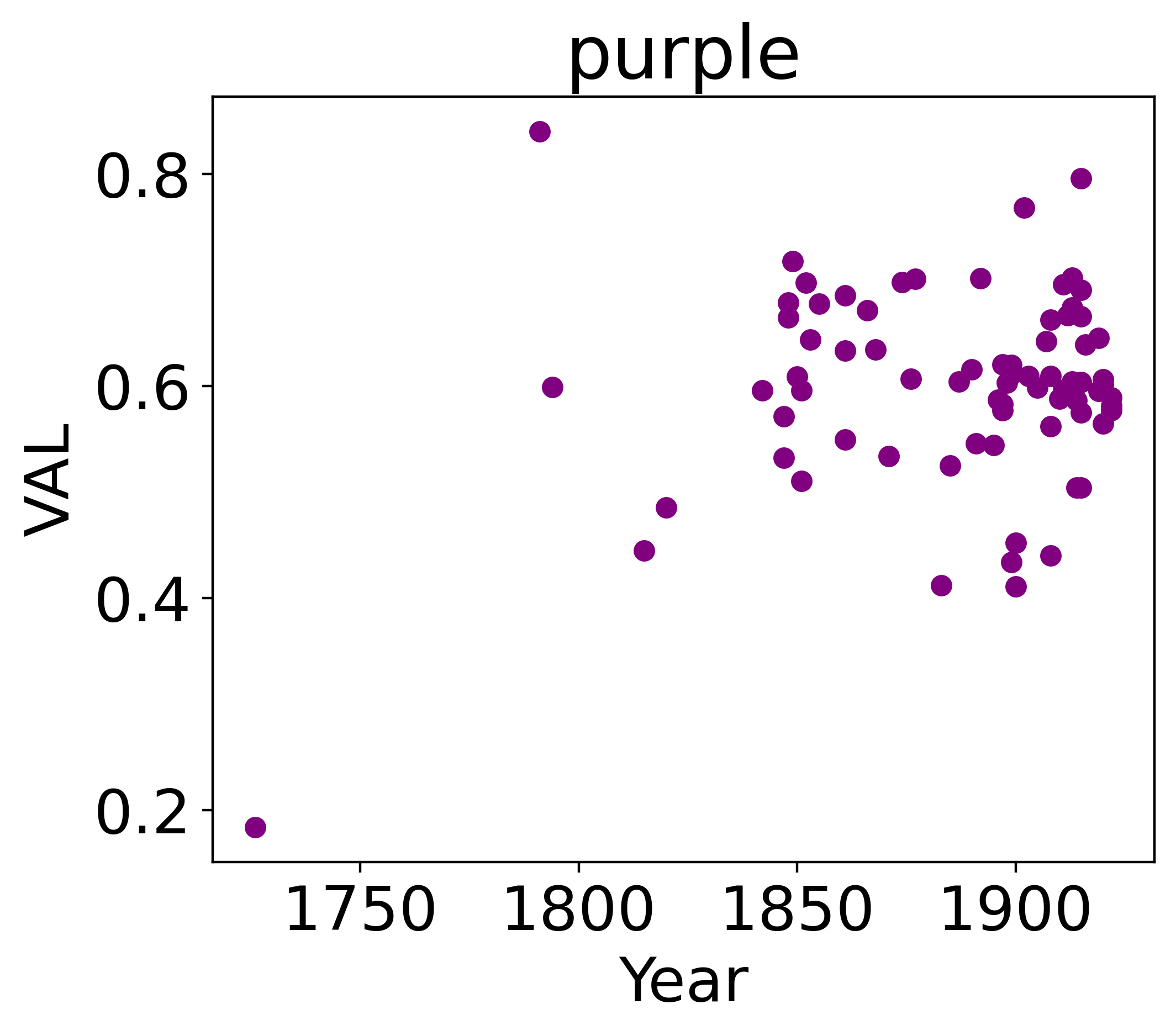}
    \caption{VAL scatter plots of nouns modified by color terms for each publication in LitBank.}
    \label{fig:val}
\end{figure*}

The plots corresponding to how color-modified nouns change over time with respect to their Glasgow Norm IMAG, CNC, and VAL values are presented here in Figures~\ref{fig:imag}, \ref{fig:cnc}, and \ref{fig:val}. We also include the full list of Pearson correlation results for all color terms in Table~\ref{tab:color_pearsons_full}.

\begin{table}[!h]
    \centering
    \begin{center}
        \textbf{IMAG Results}
    \end{center}
    \begin{tabular}{cc|cc}
    \toprule
        \textbf{Color} & \textbf{Pearson's r} & \textbf{Color} & \textbf{Pearson's r} \\
        \midrule
        red & -0.095* & green & 0.059 \\
        \midrule
        black & 0.257** & white & 0.303*** \\
        \midrule
        blue & -0.163 & brown & 0.129 \\
        \midrule
        gray & -0.081 & yellow & 0.340** \\
        \midrule
        pink & 0.534*** & purple & 0.162 \\
        \bottomrule
    \end{tabular}
    \begin{center}
        \textbf{CNC Results}
    \end{center}
    \begin{tabular}{cc|cc}
    \toprule
        \textbf{Color} & \textbf{Pearson's r} & \textbf{Color} & \textbf{Pearson's r} \\
        \midrule
        red & -0.054 & green & 0.160 \\
        \midrule
        black & 0.237** & white & 0.239** \\
        \midrule
        blue & -0.133 & brown & 0.175 \\
        \midrule
        gray & -0.044 & yellow & 0.318*** \\
        \midrule
        pink & 0.517*** & purple & 0.055 \\
        \bottomrule
    \end{tabular}
    \begin{center}
        \textbf{VAL Results}
    \end{center}
    \begin{tabular}{cc|cc}
    \toprule
        \textbf{Color} & \textbf{Pearson's r} & \textbf{Color} & \textbf{Pearson's r} \\
        \midrule
        red & -0.025 & green & 0.273*** \\
        \midrule
        black & 0.059 & white & 0.161 \\
        \midrule
        blue & -0.104 & brown & -0.148 \\
        \midrule
        gray & 0.072 & yellow & -0.01 \\
        \midrule
        pink & 0.081 & purple & 0.210* \\
        \bottomrule
    \end{tabular}
    \caption{Full list of Pearson's correlations between published years and Glasgow Norm values of the 100 fictions for each color term. *** $p < 0.001$, ** $p < 0.05$, and * $p < 0.1$.}
    \label{tab:color_pearsons_full}
\end{table}

\section{Example of Color Term Usages} \label{app:examples}
In the following usage examples, the color terms are underlined, bolded and colored, while the dependent noun is bolded. 

\subsection{Pink}
\noindent \textit{HENRY FIELDING, 1749}

Adorned with all the charms in which nature can array her; bedecked with beauty, youth, sprightliness, innocence, modesty, and tenderness, breathing sweetness from her \underline{\textbf{\color{magenta} rosy}} \textbf{lips}, and darting brightness from her sparkling eyes, the lovely Sophia comes!\\

\noindent \textit{FANNY BURNEY, 1778}

We were sitting in this manner, he conversing with all gaiety, I looking down with all foolishness, when that fop who had first asked me to dance, [...] I interrupted him I \underline{\textbf{\color{magenta} blush}} for my \textbf{folly}, with laughing; yet I could not help it; for, added to the man's stately foppishness, [...] that I could not for my life preserve my gravity.\\

\noindent \textit{ANN WARD RADCLIFFE, 1794}

As they descended, they saw at a distance, on the right, one of the grand passes of the Pyrenees into Spain, gleaming with its battlements and towers to the splendour of the setting rays, yellow tops of woods colouring the steeps below, while far above aspired the snowy points of the mountains, still reflecting a \underline{\textbf{\color{magenta} rosy}} \textbf{hue}.

Conscious innocence could not prevent a blush from stealing over Emily’s cheek; she trembled, and looked confusedly under the bold eye of Madame Cheron, who blushed also; but hers was the \underline{\textbf{\color{magenta} blush}} of \textbf{triumph}, such as sometimes stains the countenance of a person, congratulating himself on the penetration which had taught him to suspect another, and who loses both pity for the supposed criminal, and indignation of his guilt, in the gratification of his own vanity\\

\noindent \textit{CHARLOTTE BRONTE, 1847}

Her beauty, her \underline{\textbf{\color{magenta} pink}} \textbf{cheeks} and golden curls, seemed to give delight to all who looked at her, and to purchase indemnity for every fault.

Its garden, too, glowed with flowers: hollyhocks had sprung up tall as trees, lilies had opened, tulips and roses were in bloom; the borders of the little beds were gay with \underline{\textbf{\color{magenta} pink}} \textbf{thrift} and crimson double daisies; the sweetbriars gave out, morning and evening, their scent of spice and apples; and these fragrant treasures were all useless for most of the inmates of Lowood, except to furnish now and then a handful of herbs and blossoms to put in a coffin. 

She pulled out of her box, about ten minutes ago, a little \underline{\textbf{\color{magenta} pink}} silk \textbf{frock}; rapture lit her face as she unfolded it; coquetry runs in her blood, blends with her brains, and seasons the marrow of her bones. \\

\noindent \textit{ARTHUR CONAN DOYLE, 1892}

That trick of staining the fishes’ \textbf{scales} of a delicate \underline{\textbf{\color{magenta} pink}} is quite peculiar to China.

Her violet eyes shining, her lips parted, a \underline{\textbf{\color{magenta} pink}} \textbf{flush} upon her cheeks, all thought of her natural reserve lost in her overpowering excitement and concern.

As we approached, the door flew open, and a little blonde woman stood in the opening, clad in some sort of light mousseline de soie, with a touch of fluffy \underline{\textbf{\color{magenta} pink}} \textbf{chiffon} at her neck and wrists.\\

\noindent \textit{H.G. WELLS, 1897}

It was the fact that all his forehead above his blue glasses was covered by a white bandage, and that another covered his ears, leaving not a scrap of his face exposed excepting only his \underline{\textbf{\color{magenta} pink}}, peaked \textbf{nose}. \\

\noindent \textit{P. G. WODEHOUSE, 1919}

Jimmy turned into that drug store at the top of the Haymarket at which so many Londoners have found healing and comfort on the morning after, and bought the \underline{\textbf{\color{magenta} pink}} \textbf{drink} for which his system had been craving since he rose from bed.

The clerk had finished writing the ticket, and was pressing labels and a \underline{\textbf{\color{magenta} pink}} \textbf{paper} on him.\\

\noindent \textit{F. SCOTT FITZGERALD, 1920:}

\indent Myra sprang up, her cheeks \underline{\textbf{\color{magenta} pink}} with bruised \textbf{vanity}, the great bow on the back of her head trembling sympathetically.

\subsection{White}
\noindent \textit{FANNY BURNEY, 1778}

She told her niece, that she had been indulging in fanciful sorrows, and begged she would have more regard for decorum, than to let the world see that she could not renounce an improper attachment; at which Emily’s \underline{\textbf{\color{cyan} pale}} \textbf{cheek} became flushed with crimson, but it was the blush of pride, and she made no answer. \\

\noindent \textit{MARY WOLLSTONECRAFT, 1788}

Henry started at the sight of her altered appearance; the day before her complexion had been of the most \underline{\textbf{\color{cyan} pallid}} \textbf{hue}; but now her cheeks were flushed, and her eyes enlivened with a false vivacity, an unusual fire.

\noindent \textit{JANE AUSTEN, 1813}

Her \underline{\textbf{\color{cyan} pale}} \textbf{face} and impetuous manner made him start, and before he could recover himself to speak, she, in whose mind every idea was superseded by Lydia’s situation, hastily exclaimed, “I beg your pardon, but I must leave you.

From his garden, Mr. Collins would have led them round his two meadows; but the ladies, not having shoes to encounter the remains of a \underline{\textbf{\color{cyan} white}} \textbf{frost}, turned back

\noindent \textit{ELIZABETH CLEGHORN GASKELL, 1855}

She lay curled up on the sofa in the back drawing room in Harley Street looking very lovely in her \underline{\textbf{\color{cyan} white}} \textbf{muslin} and blue ribbons.

She found out that the water in the urn was cold, and ordered up the great kitchen tea kettle; the only consequence of which was that when she met it at the door, and tried to carry it in, it was too heavy for her, and she came in pouting, with a black mark on her muslin gown, and a little round \underline{\textbf{\color{cyan} white}} \textbf{hand} indented by the handle, which she took to show to Captain Lennox, just like a hurt child, and, of course, the remedy was the same in both cases.

But he had the same large, soft eyes as his daughter, eyes which moved slowly and almost grandly round in their orbits, and were well veiled by their transparent \underline{\textbf{\color{cyan} white}} \textbf{eyelids}.\\

\noindent \textit{CHARLES DICKENS, 1861}

I lighted my fire, which burnt with a raw \underline{\textbf{\color{cyan} pale}} \textbf{flare} at that time of the morning, and fell into a doze before it.

She's all in white,' he says, `wi' \underline{\textbf{\color{cyan} white}} \textbf{flowers} in her hair, and she's awful mad, and she's got a shroud hanging over her arm, and she says she'll put it on me at five in the morning.'

When we was put in the dock, I noticed first of all what a gentleman Compeyson looked, wi' his curly hair and his black clothes and his \underline{\textbf{\color{cyan} white}} pocket \textbf{handkercher}, and what a common sort of a wretch I looked. \\

\noindent \textit{ELIZABETH VON ARNIM, 1898}

There are so many bird cherries round me, great trees with branches sweeping the grass, and they are so wreathed just now with \underline{\textbf{\color{cyan} white}} \textbf{blossoms} and tenderest green that the garden looks like a wedding.

When that time came, and when, before it was over, the acacias all blossomed too, and four great clumps of \underline{\textbf{\color{cyan} pale}}, silvery pink \textbf{peonies} flowered under the south windows, I felt so absolutely happy, and blest, and thankful, and grateful, that I really cannot describe it.

\subsection{Black}

\noindent \textit{JONATHAN SWIFT, 1726}

In his right waistcoat pocket we found a prodigious bundle of white thin substances, folded one over another, about the bigness of three men, tied with a strong cable, and marked with \textbf{\underline{black}} \textbf{figures}; which we humbly conceive to be writings, every letter almost half as large as the palm of our hands.

In the left pocket were two \textbf{\underline{black}} \textbf{pillars} irregularly shaped: we could not, without difficulty, reach the top of them, as we stood at the bottom of his pocket.

In one of these cells were several globes, or balls, of a most ponderous metal, about the bigness of our heads, and requiring a strong hand to lift them: the other cell contained a heap of certain \textbf{\underline{black}} \textbf{grains}, but of no great bulk or weight, for we could hold above fifty of them in the palms of our hands.

About two or three days before I was set at liberty, as I was entertaining the court with this kind of feat, there arrived an express to inform his majesty, that some of his subjects, riding near the place where I was first taken up, had seen a great \textbf{\underline{black}} \textbf{substance} lying on the ground, [...] they would undertake to bring it with only five horses. \\

\noindent \textit{FANNY BURNEY, 1778}

You can't think how oddly my head feels; full of powder and \textbf{\underline{black}} \textbf{pins}, and a great cushion on the top of it.

if you'll say that, you'll say anything: however, if you swear till you're \textbf{\underline{black}} in the \textbf{face}, I sha'n't believe you; for nobody sha'n't persuade me out of my senses, that I'm resolved.

Ridiculous, I told him, was a term which he would find no one else do him the favour to make use of, in speaking of the horrible actions belonging to the old story he made so light of; 'actions' continued I, 'which would dye still deeper the \textbf{\underline{black}} \textbf{annals} of Nero or Caligula. \\

\noindent \textit{JANE AUSTEN, 1813}

The ladies were somewhat more fortunate, for they had the advantage of ascertaining from an upper window, that he wore a blue coat and rode a \textbf{\underline{black}} \textbf{horse}. \\

\noindent \textit{CHARLOTTE BRONTE, 1847}

And then she had such a fine \textbf{head} of hair; raven \textbf{\underline{black}} and so becomingly arranged: a crown of thick plaits behind, and in front the longest, the glossiest curls I ever saw.

Afternoon arrived: Mrs. Fairfax assumed her best \textbf{\underline{black}} satin \textbf{gown}, her gloves, and her gold watch; for it was her part to receive the company,- to conduct the ladies to their rooms, Adele, too, would be dressed: though I thought she had little chance of being introduced to the party that day at least.

Fluttering veils and waving plumes filled the vehicles; two of the cavaliers were young, dashing looking gentlemen; the third was Mr. Rochester, on his \textbf{\underline{black}} \textbf{horse}, Mesrour, Pilot bounding before him; at his side rode a lady, and he and she were the first of the party.

The noble bust, the sloping shoulders, the graceful neck, the dark eyes and \textbf{\underline{black}} \textbf{ringlets} were all there; - but her face?

Mrs. Fairfax was summoned to give information respecting the resources of the house in shawls, dresses, draperies of any kind; and certain wardrobes of the third storey were ransacked, and their contents, in the shape of brocaded and hooped petticoats, satin sacques, \textbf{\underline{black}} \textbf{modes}, lace lappets, etc, were brought down in armfuls by the abigails; then a selection was made, and such things as were chosen were carried to the boudoir within the drawing room. \\

\noindent \textit{NATHANIEL HAWTHORNE, 1850}

Doubtless, however, either of these stern and \textbf{\underline{black}} browed \textbf{Puritans} would have thought it quite a sufficient retribution for his sins that, after so long a lapse of years, the old trunk of the family tree, with so much venerable moss upon it, should have borne, as its topmost bough, an idler like myself.

The Custom House marker imprinted it, with a stencil and \textbf{\underline{black}} \textbf{paint}, on pepper bags, and baskets of anatto, and cigar boxes, and bales of all kinds of dutiable merchandise, in testimony that these commodities had paid the impost, and gone regularly through the office.

Before this ugly edifice, and between it and the wheel track of the street, was a grass plot, much overgrown with burdock, pig weed, apple pern, and such unsightly vegetation, which evidently found something congenial in the soil that had so early borne the \textbf{\underline{black}} \textbf{flower} of civilised society, a prison. \\

\noindent \textit{CHARLES DICKENS, 1861}

I still held her forcibly down with all my strength, like a prisoner who might escape; and I doubt if I even knew who she was, or why we had struggled, or that she had been in flames, or that the flames were out, until I saw the patches of tinder that had been her garments no longer alight but falling in a \textbf{\underline{black}} \textbf{shower} around us.

The sudden exclusion of the night, and the substitution of \textbf{\underline{black}} \textbf{darkness} in its place, warned me that the man had closed a shutter.

He had a boat cloak with him, and a \textbf{\underline{black}} canvas \textbf{bag}; and he looked as like a river pilot as my heart could have wished.

The marshes were just a long \textbf{\underline{black}} horizontal \textbf{line} then, as I stopped to look after him; and the river was just another horizontal line, not nearly so broad nor yet so black; and the sky was just a row of long angry red lines and dense black lines intermixed. \\

\noindent \textit{G.K. CHESTERTON, 1908}

The tall hat and long frock \textbf{coat} were \textbf{\underline{black}}; the face, in an abrupt shadow, was almost as dark.

He wore an old fashioned \textbf{\underline{black}} chimney pot \textbf{hat}; he was wrapped in a yet more old fashioned \textbf{cloak}, \textbf{\underline{black}} and ragged; and the combination gave him the look of the early villains in Dickens and Bulwer Lytton.

A long, lean, \textbf{\underline{black}} \textbf{cigar}, bought in Soho for twopence, stood out from between his tightened teeth, and altogether he looked a very satisfactory specimen of the anarchists upon whom he had vowed a holy war.

He had a \textbf{\underline{black}} French \textbf{beard} cut square and a \textbf{\underline{black}} English frock \textbf{coat} cut even squarer. \\

\noindent \textit{SOMERSET W. MAUGHAM, 1915}

She wore a \textbf{\underline{black}} \textbf{dress}, and her only ornament was a gold chain, from which hung a cross.

It was a large \textbf{\underline{black}} \textbf{stove} that stood in the hall and was only lighted if the weather was very bad and the Vicar had a cold.

And the poor lady, so small in her \textbf{\underline{black}} \textbf{satin}, shrivelled up and sallow, with her funny corkscrew curls, took the little boy on her lap and put her arms around him and wept as though her heart would break.

\subsection{Yellow}

\noindent \textit{DANIEL DEFOE, 1719}

The colour of his skin was not quite black, but very tawny; and yet not an ugly, \underline{\textbf{\color{yellow} yellow}}, nauseous \textbf{tawny}, as the Brazilians and Virginians, and other natives of America are, but of a bright kind of a dun olive colour, that had in it something very agreeable, though not very easy to describe.

\noindent \textit{JONATHAN SWIFT, 1726}

The projector of this cell was the most ancient student of the academy; his face and \textbf{beard} were of a pale \underline{\textbf{\color{yellow} yellow}}; his hands and clothes daubed over with filth.

The hair of both sexes was of several \textbf{colours}, brown, red, black, and \underline{\textbf{\color{yellow} yellow}}.

I forgot another circumstance (and perhaps I might have the reader’s pardon if it were wholly omitted), that while I held the odious vermin in my hands, it voided its filthy excrements of a \underline{\textbf{\color{yellow} yellow}} liquid \textbf{substance} all over my clothes; but by good fortune there was a small brook hard by, where I washed myself as clean as I could; although I durst not come into my master’s presence until I were sufficiently aired. \\

\noindent \textit{WILLIAM MAKEPEACE THACKERAY, 1848}

Joseph still continued a huge clattering at the poker and tongs, puffing and blowing the while, and turning as red as his \underline{\textbf{\color{yellow} yellow}} \textbf{face} would allow him.

Why, he had the \underline{\textbf{\color{yellow} yellow}} \textbf{fever} three times; twice at Nassau, and once at St.

At one end of the hall is the great staircase all in black oak, as dismal as may be, and on either side are tall doors with stags' heads over them, leading to the billiard room and the library, and the great \underline{\textbf{\color{yellow} yellow}} \textbf{saloon} and the morning rooms.

He's never content unless he gets my \underline{\textbf{\color{yellow} yellow}} sealed \textbf{wine}, which costs me ten shillings a bottle, hang him! \\

\noindent \textit{JAMES JOYCE, 1914}

I liked the last best because its \textbf{leaves} were \underline{\textbf{\color{yellow} yellow}}.

Breakfast was over in the boarding house and the table of the breakfast room was covered with plates on which lay \underline{\textbf{\color{yellow} yellow}} \textbf{streaks} of eggs with morsels of bacon fat and bacon rind.

An immense scarf of peacock blue muslin was wound round her hat and knotted in a great bow under her chin; and she wore bright \underline{\textbf{\color{yellow} yellow}} \textbf{gloves}, reaching to the elbow.

His face, shining with raindrops, had the appearance of damp \underline{\textbf{\color{yellow} yellow}} \textbf{cheese} save where two rosy spots indicated the cheekbones.

While the point was being debated a tall agile gentleman of fair complexion, wearing a long \underline{\textbf{\color{yellow} yellow}} \textbf{ulster}, came from the far end of the bar.

I saw that he had great gaps in his mouth between his \underline{\textbf{\color{yellow} yellow}} \textbf{teeth}.

On the closed square piano a pudding in a huge \underline{\textbf{\color{yellow} yellow}} \textbf{dish} lay in waiting and behind it were three squads of bottles of stout and ale and minerals, drawn up according to the colours of their uniforms, the first two black, with brown and red labels, the third and smallest squad white, with transverse green sashes.

A dull \underline{\textbf{\color{yellow} yellow}} light brooded over the houses and the river; and the sky seemed to be descending. \\

\noindent \textit{F. SCOTT FITZGERALD, 1920}

The Gothic halls and cloisters were infinitely more mysterious as they loomed suddenly out of the darkness, outlined each by myriad faint squares of \underline{\textbf{\color{yellow} yellow}} \textbf{light}.

His face was cast in the same \underline{\textbf{\color{yellow} yellow}} \textbf{wax} as in the cafe, neither the dull, pasty color of a dead man—rather a sort of virile pallor—nor unhealthy, you’d have called it; but like a strong man who’d worked in a mine or done night shifts in a damp climate.

The two hours’ ride were like days, and he nearly cried aloud with joy when the towers of Princeton loomed up beside him and the \underline{\textbf{\color{yellow} yellow}} \textbf{squares} of light filtered through the blue rain.

Browsing in her library, Amory found a tattered gray book out of which fell a \underline{\textbf{\color{yellow} yellow}} \textbf{sheet} that he impudently opened.

There was that shade of glorious \underline{\textbf{\color{yellow} yellow}} \textbf{hair}, the desire to imitate which supports the dye industry.\\

\noindent \textit{F. SCOTT FITZGERALD, 1922}

That this feeble, unintelligent old man was possessed of such power that, \underline{\textbf{\color{yellow} yellow}} \textbf{journals} to the contrary, the men in the republic whose souls he could not have bought directly or indirectly would scarcely have populated White Plains, seemed as impossible to believe as that he had once been a pink and white baby.

He bulges in other places his paunch bulges, prophetically, his words have an air of bulging from his mouth, even his dinner coat pockets bulge, as though from contamination, with a dog eared collection of time tables, programmes, and miscellaneous scraps on these he takes his notes with great screwings up of his unmatched \underline{\textbf{\color{yellow} yellow}} \textbf{eyes} and motions of silence with his disengaged left hand.

It was a crackling dusk when they turned in under the white façade of the Plaza and tasted slowly the \textbf{foam} and \underline{\textbf{\color{yellow} yellow}} thickness of an egg nog.

He had fixed his aunt with the bright \underline{\textbf{\color{yellow} yellow}} \textbf{eye}, giving her that acute and exaggerated attention that young males are accustomed to render to all females who are of no further value.

On the appointed Wednesday in February Anthony had gone to the imposing offices of Wilson, Hiemer and Hardy and listened to many vague instructions delivered by an energetic young man of about his own age, named Kahler, who wore a defiant \underline{\textbf{\color{yellow} yellow}} \textbf{pompadour}, and in announcing himself as an assistant secretary gave the impression that it was a tribute to exceptional ability.

Joe Hull had a \underline{\textbf{\color{yellow} yellow}} \textbf{beard} continually fighting through his skin and a low voice which varied between basso profundo and a husky whisper.

\end{document}